\documentclass{article} 
\usepackage[margin=1in]{geometry}
\usepackage[utf8]{inputenc}
\usepackage[noblocks]{authblk}
\usepackage{hyperref}
\usepackage{subfigure}
\usepackage{verbatim}
\usepackage{nicefrac}       
\usepackage{microtype}      
\usepackage{pifont}
\usepackage{natbib}
\usepackage{graphicx}
\graphicspath{{../}} 
\usepackage{amssymb}
\usepackage{amsmath}
\usepackage{amsthm}
\usepackage{epstopdf}
\usepackage{color}
\usepackage{enumitem}

\usepackage{multirow}
\usepackage{sidecap}
\usepackage{xfrac}
\usepackage{booktabs}
\usepackage{wrapfig}

\newcommand\blfootnote[1]{%
	\begingroup
	\renewcommand\thefootnote{}\footnote{#1}%
	\addtocounter{footnote}{-1}%
	\endgroup
}

\newtheorem{theorem}{Theorem}
\newtheorem{definition}{Definition}

\newtheorem{lemma}{Lemma}
\newtheorem{proposition}{Proposition}

\newcommand{\bs}[1]{\boldsymbol{#1}}
\newcommand{\bsb}{\bs{w}}

\newcommand{\E}[1]{\mathbb{E}\left[#1\right]}

\begin{document}
\title{On Generalization of Adaptive Methods for Over-parameterized Linear Regression}

\author[1]{Vatsal Shah*}
\author[1]{Soumya Basu*}
\author[2]{Anastasios Kyrillidis}
\author[1]{Sujay Sanghavi}
\affil[1]{Department of Electrical and Computer Engineering, UT Austin}
\affil[2]{Computer Science Department, Rice University}

\renewcommand\Authands{ and }

\maketitle

\begin{abstract}
Over-parameterization and adaptive methods have played a crucial role in the success of deep learning in the last decade. The widespread use of over-parameterization has forced us to rethink generalization by bringing forth new phenomena, such as implicit regularization of optimization algorithms and double descent with training progression. A series of recent works have started to shed light on these areas in the quest to understand -- \textit{why do neural networks generalize well?} The setting of over-parameterized linear regression has provided key insights into understanding this mysterious behavior of neural networks. 

In this paper, we aim to characterize the performance of adaptive methods in the over-parameterized linear regression setting. First, we focus on two sub-classes of adaptive methods depending on their generalization performance. For the first class of adaptive methods, the parameter vector remains in the span of the data and converges to the minimum norm solution like gradient descent (GD).  On the other hand, for the second class of adaptive methods, the gradient rotation caused by the pre-conditioner matrix results in an in-span component of the parameter vector that converges to the minimum norm solution and the out-of-span component that saturates. Our experiments on over-parameterized linear regression and deep neural networks support this theory.  
%
\end{abstract}

\section{Introduction}\blfootnote{* indicates equal contribution}
\blfootnote{Corresponding Emails: vatsalshah1106@utexas.edu, basusoumya@utexas.edu, anastasios@rice.edu, sanghavi@mail.utexas.edu}

The success of deep learning has uncovered a new mystery of \textit{benign overfitting} \citep{bartlett2020benign, muthukumar2020harmless}, i.e., systems with a large number of parameters can not only achieve zero training error but are also able to generalize well. Also, over-parameterized systems exhibit a double descent-behavior \citep{bartlett2020benign, belkin2018understand}; \textit{as the number of parameters/epochs increases, the test error first decreases, then increases before falling again}. This goes against the conventional wisdom of overfitting in machine learning, which stems from the classical bias-variance tradeoff \citep{bishop2006pattern,friedman2001elements,shalev2014understanding}. 

In the absence of explicit regularization, a typical over-parameterized setting possesses multiple global minima. Classical gradient descent based methods can achieve one of these many global minima \citep{allen2018convergence, du2018gradient, du2018power, gunasekar2017implicit, soudry2018implicit}, however not all optima generalize equally. \citep{bengio2012practical, bottou2012stochastic, orr2003neural} suggest many practical approaches to improve generalization; however, there remains a considerable gap between theory and practice \citep{zhang2016understanding, allen2018convergence}.

In this paper, we will focus on two categories of optimization algorithms: pure gradient descent based (non-adaptive\footnote{Now onwards, optimization methods that satisfy equation \eqref{eq:nonadaptive} will be referred to as non-adaptive purposes.}) methods and adaptive methods. The primary distinguishing factor between these two methods is determined by the update step. For the class of non-adaptive methods, the expected gradient update step is given as follows: 
\begin{align} 
\E{\bs{w}(t+1)|\bs{w}(t)} = \bs{w}(t) - \eta \nabla f(\bs{w}(t)),\label{eq:nonadaptive}
\end{align}
where $\bs{w}(t)$ indicates the estimate of the underlying parameter vector, $\eta$ represents the learning rate and $f(\bs{w}(t)), \nabla f(\bs{w}(t))$ represent the loss function and its gradient, respectively. Popular methods like gradient descent, stochastic gradient descent (SGD), batch gradient descent fall under this class.
Training any model using non-adaptive methods involves tuning over many hyperparameters, of which \emph{step size} is the most essential one \citep{sutskever2013importance, schaul2013no}.
The step size could be set as constant, or could be changing per iteration $\eta(t)$ \citep{bottou2012stochastic}, usually based on a predefined learning rate schedule \citep{bottou2010large, xu2011towards, senior2013empirical}.

During the past decade, we have also witnessed the rise of a family of algorithms called adaptive methods that argue for \emph{automatic} hyper-parameter adaptation \citep{ruder2016overview} during training (including step size).
The list includes AdaGrad \citep{duchi2011adaptive}, Adam \citep{kingma2014adam}, AdaDelta \citep{zeiler2012adadelta}, RMSProp \citep{tieleman2012lecture}, AdaMax \citep{kingma2014adam}, Nadam \citep{dozat2016incorporating}, just to name a few.
These algorithms utilize current and past gradient information $\{\nabla f(\bs{w}(i))\}_{i = t}^k$, for $t < k$, to design preconditioning matrices $\bs{D}(t) \succeq 0$ that better pinpoint the local curvature of the objective function as follows:
\begin{align} 
\E{\bs{w}(t+1)|\bs{w}(t)} = \bs{w}(t) - \eta \bs{D}(t) \nabla f(\bs{w}(t))\label{eq:adaptive}
\end{align}
Usually, the main argument for using adaptive methods is that $\bs{D}(t)$ eliminates pre-setting a learning rate schedule, or diminishes initial bad step size choices, thus, detaching the time-consuming part of step size tuning from the practitioner \citep{zhang2017yellowfin}. 

\citep{gunasekar2018characterizing} was one of the first papers to discuss the implicit bias introduced by optimization methods for over-parameterized systems and how the choice of optimization algorithm affects the global minima it attains. However, the generalization behavior of these optimization methods remains a mystery. As a result, researchers have re-focussed their attention on understanding the most straightforward over-parameterized setting of linear regression \citep{bartlett2020benign,hastie2019surprises,mei2019generalization,muthukumar2020harmless} as a first step in unraveling the mysterious behavior of neural networks. 

Gradient descent-based methods converge to the minimum norm interpolated solution \citep{muthukumar2020harmless} for over-parameterized linear regression. Under certain assumptions on the data distribution, the minimum norm solution achieves near-optimal accuracy for unseen data \citep{bartlett2020benign}. Unlike SGD, the presence of $\bs{D}(t)$ in adaptive methods can alter the span of the final converged solution in the presence of any non-trivial initialization, which makes the task of commenting on adaptive methods challenging.

Despite being a key reason behind the success of deep learning, the convergence behavior of adaptive methods is not well understood. The convergence bounds for most adaptive methods hold for only a specific pre-conditioner matrix \citep{kingma2014adam, tran2019convergence, gunasekar2018characterizing}. Besides, theoretical guarantees for adaptive methods often minimize regret \citep{reddi2019convergence, duchi2011adaptive}, which makes it further challenging to comment on the generalization of adaptive methods.
As a result, the generalization of adaptive methods for a general $\bs{D}(t)$ remains an open problem even for an over-parameterized linear regression setting.
In this paper, we aim to explicitly characterize the sub-class of adaptive methods that mimic the \textit{convergence}, and \textit{generalization} behaviors seen in SGD and the sub-class that does not. In addition, we observe a double descent like phenomena for a sub-class of adaptive methods as the number of training epochs increases.

\begin{table*}[!t]
	\centering
	\vspace{0.3cm}
	\begin{footnotesize}
		\begin{tabular}{c c c c c c c}
			\toprule 
			$d=50$  & \phantom{5} & GD &  AM1 & AM2  &AM3\\  \midrule
			\multirow{3}{*}{$n = 10$} & Training Error &$1.27\cdot 10^{-28}$ &$\mathbf{1.42\cdot 10^{-29}}$ &$8.64\cdot 10^{-4}$ &$8.64\cdot 10^{-29}$ \\
			\phantom{1} & Test Error  &$81.56$ &$\mathbf{76.94}$ &$79.62$ &$81.65$\\
			\phantom{1} & $\|\bs{w} - \bs{w}^*\|$ &$9.08$ &$\mathbf{8.92}$ &$9.03$ &$9.08$\\
			\midrule
			\multirow{3}{*}{$n = 40$} & Training Error &$4.77\cdot 10^{-5}$ &$\mathbf{6.07\cdot 10^{-7}}$ &$3.31\cdot 10^{-3}$ &$8.64\cdot 1.17^{-4}$ \\
			\phantom{1} & Test Error &$\mathbf{18.62}$ &$19.56$  &$20.35$ &$18.65$ \\
			\phantom{1} & $\|\bs{w} - \bs{w}^*\|$ &$\mathbf{4.31}$ &$4.37$ &$4.51$ &$\mathbf{4.31}$\\
			\bottomrule
		\end{tabular}
	\end{footnotesize}\caption{Table illustrating differing generalization guarantees of three distinct Adaptive Methods (AM) with SGD in overparameterized setting, i.e. $d>n$, where $n$: number of examples, $d$: dimension. AM1: Diagonalized Adagrad, AM2: Adagrad (AM1) Variant (where we square the diagonal terms instead of taking the square root), AM3: Projected version of AM1 onto the span of $\bs{X}$. The exact expressions for the pre-conditioner matrix are available in Section 3.}    \label{table:motivation}
\end{table*}

In this paper, we would like to understand \textit{how adaptive methods affect generalization guarantees of over-parameterized problems}. To motivate this, we consider a toy example for simple linear regression in the under-determined/over-parameterized framework in Table \ref{table:motivation}. 
As is evident, some adaptive methods have the same generalization as SGD, while others can yield quite different generalization guarantees. \\
\textbf{Key Contributions:}
For the \textit{theoretical contribution}, we focus on over-parameterized linear regression. 
Here, plain gradient descent methods converge to the \emph{minimum Euclidean norm solution}, while adaptive methods may or may not. 
In this paper, we provide explicit conditions on the structure of pre-conditioner matrices, $\bs{D}(t)$, which allow us to distinguish between two classes of adaptive methods, the ones which behave similarly to SGD and the ones that do not.
Based on these conditions, we compare the generalization performance between adaptive and non-adaptive methods. 

For the \textit{experimental component}, we begin by revisiting the mystery posed by Table \ref{table:motivation}, and demonstrate that the experimental results are in line with our theoretical guarantees. 
Further, we show using a toy example that the adaptive methods can have a superior generalization performance than SGD.
The discussion \textit{``which method is provably better''}, however, is inconclusive and ultimately depends on the problem/application at hand. 
Lastly, we empirically demonstrate the validity of our claims for over-parameterized neural networks as well
and recommend exercising caution when proposing or choosing adaptive methods for training, depending on the goal in hand.

\section{Problem Setup}
\noindent \textbf{Notation.} For any matrix $\bs{A} \in \mathbb{R}^{m\times n}$, $A_{pq}$ indicates the element corresponding to the $p$-th row and $q$-th column. 
The $\text{rank}(\bs{A})$ denotes the rank of $\bs{A}$. 
For a sequence of matrices $\bs{A}_0$ to $\bs{A}_n$, we have the definition $\prod_{k=i+m}^{i} \bs{A}_k = \bs{A}_{(i+m)} \bs{A}_{(i+m-1)} \dots \bs{A}_{i}$. 
Note that, $a(t)$ indicates the value of the the function $a(\cdot)$ after the $t$-th update. 
Note that $\lambda$ without any subscript indicates the regularizer, and $\lambda_i$ with a subscript denotes the $i^{th}$ eigenvalue. The subscript $\bs{.}_{(1)}$ will denote the in-span component, subscript $\bs{.}_{(2)}$ will denote the out-of-span component and $\bs{.}(t)$ will indicate the $t^{th}$ iterate.

We consider an over-parameterized noisy linear regression (possibly with regularization), where the relationship between the data matrix $\bs{X} \in \mathbb{R}^{n\times d}$, the noise vector $\bs{\zeta} \in \mathbb{R}^{n}$, and the labels $\bs{y}\in \mathbb{R}^{n}$ is as follows: 
\begin{align}
\bs{y} = \bs{X} \bsb^\star + \bs{\zeta}.
\end{align}
We are concerned with the following optimization problem:
\begin{align}
f(\bsb) = \arg \min_{\bsb} \left\{\mathbb{E}\left[\|\bs{y} - \bs{X}\bsb\|^2\right] + \dfrac{\lambda}{2} \|\bsb\|_2^2\right\}.
\end{align}
In particular, we study the convergence of the following iterative updates
\begin{equation}\label{eq:linUpdate}
\bsb(t+1) = \bsb(t) - \eta \bs{D}(t) \nabla f(\bsb(t)),
\end{equation}
where the pre-conditioner matrices are \emph{bounded}, \emph{positive definite} and hence full rank; i.e., $\inf_t \text{rank}(\bs{D}(t)) = d$.
The system is assumed over-parameterized; i.e., $R = \text{rank}(\bs{X}) < d$. 

Before we discuss the generalization of adaptive methods in over-parameterized settings, let us briefly explain their performance on the training set. For linear regression with $\ell_2$-norm regularization, we observe that adaptive methods with any full rank pre-conditioner matrix $\bs{D}(t)$ will converge to the same solution as its non-adaptive counterpart and thus mimic their performance. However, for unregularized linear regression, adaptive methods can converge to entirely different solutions than SGD. 
Both SGD and adaptive methods can achieve zero training error despite attaining different stationary points\footnote{For more, refer to the Appendix Sections A.2 and A.3.}.

\subsection{Performance on Unseen Data}
As a result, our primary focus in this paper is to understand the generalization capabilities of adaptive methods. We observe that the generalization depends on two key factors: \textit{ $i)$ Does $\bsb^\star$ lie in the span of the data matrix, $\bs{X}$? $ii)$ How does pre-multiplying with the pre-conditioner matrix alter the span of final converged $\bsb$? }

\subsubsection{Spectral Representation}
The switch to the spectral domain allows us to simplify and understand the relationship between the final converged solution with the span of data matrix $\bs{X}$, pre-conditioner matrix $\tilde{\bs{D}}(t)$ and the initialization $w(0)$.
We  express the data matrix using its singular value decomposition (SVD): $\bs{X} = \sum_{r= 1}^{R} \lambda_r \bs{u}_r \bs{v}_r^T$, $\lambda_r \neq 0$ for all $r$ where $\lambda_r, \bs{u}_r, \bs{v_r}$ represent the $r^{th}$ largest eigenvalue and the corresponding right and left eigenvectors respectively. We complete the basis using the left eigenvectors of the data matrix to form a complete orthogonal spectral basis of $\mathbb{R}^d$, $\{ \bs{v}_r: r= 1\dots, d\}$ form the basis vectors and denote it by $\bs{V}$. Similarly, $\bs{U}$ forms the complete orthogonal spectral basis of $\mathbb{R}^n$ using the right eigenvectors of the data matrix as $\{ \bs{u}_r: r= 1\dots, n\}$. The eigenvalue matrix is $\bs{\Lambda}$ where $\bs{\Lambda}_{rr} = \lambda_r$ if $1\leq r \leq R$ and $0$ otherwise.
We next express useful quantities in the above bases in Table \ref{table:notation}.
\aboverulesep=0ex
\belowrulesep=0ex


The definition of adaptive pre-conditioner matrices in the above table holds since $\bs{V}$ represents a complete orthogonal spectral basis of $\mathbb{R}^d$. Additionally, we also have the following property, where we show that pre- and post-multiplication by an orthogonal matrix $\bs{V}$ does not alter the eigenvalues of the original matrix, i.e., the set of eigenvalues for $\tilde{\bs{D}}(t)$ is identical to the set of eigenvalues of $\bs{D}(t)$ (Appendix Section A.5). 

\begin{table*}[!h]
	\centering
	\begin{tabular}{|c|c|}\noalign{\hrule height 1.0pt} 
		Data matrix &$\bs{X} = \bs{U} \bs{\Lambda} \bs{V}^T$ \\
		\noalign{\hrule height 0.5pt} 
		True parameter &$\bsb^\star = \bs{V} \tilde{\bsb}^\star$\\
		\noalign{\hrule height 0.5pt} 
		Noise vector &$\bs{\zeta} = \bs{U} \tilde{\bs{\zeta}}$ \\
		\noalign{\hrule height 0.5pt} 
		\multirow{2}{*}{Adaptive pre-conditioner matrices}
		&${\bs{D}(t) = \sum_{r=1}^{d} \sum_{s=1}^{d} \tilde{D}_{rs}(t)  \bs{v}_r \bs{v}_s^T}$,  \\        
		&$\tilde{\bs{D}}(t) = \bs{V}^T \bs{D}(t) \bs{V}, \tilde{\bs{D}}(t) \in \mathbb{R}^{d \times d}$\\
		\noalign{\hrule height 0.5pt} 
		Weight vectors &$\bsb(t) = \bs{V} \tilde{\bsb}(t)$\\
		\noalign{\hrule height 1.0pt} 
	\end{tabular}  
	\vspace{0.1cm}
	\caption{Notation in spectral domain} \label{table:notation}
\end{table*}

\subsubsection{Closed Form Expression for the Iterates}
Our objective is to understand how the iterates evolve depending on the space spanned by the data matrix.  
First, we establish a closed-form expression for the updates of the vector $\tilde{\bsb}(t)$. 

\begin{proposition}\label{lem:spectralclosed}
	Consider the over-parameterized linear regression setting with data matrix $\bs{X}$, noise $\bs{\zeta}$, and regularizer $\lambda >0$.
	If the pre-conditioner matrix $\bs{D}(t)\succ 0$ for all $t\geq 0$, then, for any $T \geq 0$, the iterate $\tilde{\bs{w}}(T)$ admits the following closed form expression: 
	\begin{small}
		\begin{align}\label{eq:spectralClosedForm}
		&\tilde{\bs{w}}(T) = \prod_{i=0}^{T-1} \left(\bs{I} - \eta \tilde{\bs{D}}(i) (\bs{\Lambda}^2 + \lambda \bs{I})\right) \tilde{\bsb}(0) \\ 
		&+
		\sum_{i= 0}^{T-1} \prod_{j=(i+1)}^{T-1} \left(\bs{I} - \eta \tilde{\bs{D}}(j) (\bs{\Lambda}^2 + \lambda \bs{I})\right)  \eta \tilde{\bs{D}}(i)(\bs{\Lambda}^2 \tilde{\bsb}^\star +\bs{\Lambda} \bs{\zeta}) \notag
		\end{align} 
	\end{small}
\end{proposition}

\noindent
The final expression of $\bs{w}(T)$ implies that the final solution depends on the initialization point, the span of the data matrix in $\mathbb{R}^d$ space, and the pre-conditioner matrix.
Further, the closed-form indicates that the presence of pre-conditioning matrices $\tilde{\bs{D}}(j)$ may cause $\bs{w}(t)$ to lie outside of the span of the data in the complete $\mathbb{R}^d$ space. 

We observe that the presence or absence of regularizer can significantly alter the stationary points to which adaptive methods converge. In the presence of $\ell_2$-norm regularization, we observe that the adaptive methods converge to the same solution independent of the initialization or the step-size. However, in the absence of regularization, things are not as straight-forward. 
Here, we will try to capture the convergence of over parameterized linear regression using dynamics described by equation \eqref{eq:linUpdate}.

\noindent
\paragraph{$\ell_2$-norm Regularized Linear Regression.} 
In presence of $\ell_2$-norm regularization, the over-parameterized linear regression problem becomes strongly convex and possesses a unique global optima. Proposition 2 serves a sanity check; where we show the convergence to this unique optima for any positive definite pre-conditioner matrix $\bs{D}(t)$ in the spectral domain. We utilize this result as a stepping stone in understanding the convergence behavior of adaptive methods for the unregularized over-parameterized settings.
\begin{proposition}\label{lem:spectralRegularized}
	Consider the over-parameterized linear regression setting with data matrix $\bs{X}$, noise $\bs{\zeta}$, and regularizer $\lambda >0$. Suppose for all $t\geq 0$, the pre-conditioner matrix $\bs{D}(t) \succ 0$, and the learning rate satisfies $$\eta \in \left(0,  2\left(\lambda_{\max}(\bs{D}(t))(\lambda_{\max}^2 (\bs{X})+ \lambda)\right)^{-1}\right).$$ where $\lambda_{\max}(\cdot)$ indicates the maximum eigenvalue.
	Then, $\tilde{\bs{w}}(t)$ converges to the following fixed point
	\begin{align}\label{eq:spectralClosedForm}
	&\lim_{t\to \infty}\tilde{\bs{w}}(t) =
	(\bs{\Lambda}^2 + \lambda \bs{I})^{-1} (\bs{\Lambda}^2 \tilde{\bsb}^\star +\bs{\Lambda} \bs{\zeta}).
	\end{align} 
\end{proposition}
Proposition \ref{lem:spectralRegularized} states that like the gradient descent based methods, the adaptive methods will perfectly capture the component of the generative $\bsb^*$ that lies in the subspace formed by the data matrix. In other words, with regularization the parameter vector converges in the span of $\bs{X}$.

In the proof of the proposition presented in the Appendix, we use contraction properties to show convergence where $\lambda > 0$ plays a significant role. 
Further, as $\inf_t \text{rank}(\bs{D}(t))$, a simple fixed-point analysis provides us with the in-span component and shows that for $\lambda > 0$ there is no out-of-span component of the solution. Note that this proposition acts as a proof of concept for the well-known result that adaptive methods and non-adaptive methods converge to the same solution in the presence of $\ell_2$-norm regularization.

Lastly, different regularization techniques alter the implicit bias of the final converged solution differently. The claims made in this sub-section are only valid for $\ell_2$-norm regularization.

\noindent
\paragraph{Unregularized Linear Regression.} 
Next, we focus on the slightly more interesting problem of unregularized linear regression in the over-parameterized regime. The optimization problem with squared loss is no longer strongly convex, and there are infinite solutions that can achieve zero training error.  In this case, the convergence of unregularized linear regression depends on the initialization $\tilde{\bs{w}}(0)$. Further, as $\lambda_{\min}(\tilde{\bs{D}}(t) \bs{\Lambda}^2) = 0$ we cannot directly prove (using contraction mapping) convergence for general unregularized over-parameterized linear regression. However, when the pre-conditioner matrices satisfy a block matrix structure, then we can say something about the converged solution. Note that most of the popular adaptive algorithms \citep{duchi2011adaptive,kingma2014adam,reddi2019convergence,wu2019global} satisfy the block matrix structure.

The out-of-span component behavior depends subtly on the interplay of the pre-conditioner matrices and the span of data. Now, we establish sufficient conditions on the class of pre-conditioner matrices for which the convergence is guaranteed (for more details refer Appendix). 

We define some notations useful to state our main theorems. 
\textit{We use $\tilde{\bs{w}}_{(1)}(\infty) = \lim\limits_{t\to \infty}\tilde{\bs{w}}_{(1)}(t)$ to denote the in-span component, and $\tilde{\bs{w}}_{(2)}(\infty) = \lim\limits_{t\to \infty}\tilde{\bs{w}}_{(2)}(t)$ to denote the in-span component of the stationary point.}\footnote{Note that $\tilde{\bs{w}}_{(2)}(\infty) \in (\mathbb{R}\cup \{\infty\})^d$ for $i=1,2$, as we can not assume convergence of the iterates a pirori.} 
Let, $e_{(1)}(t) = \|\tilde{\bs{w}}_{(1)}(\infty)  - \tilde{\bs{w}}_{(1)}(i)\|_2 = \mathcal{O}\left(\dfrac{1}{t^\beta}\right)$ be the $\ell_2$-norm distance of in-span component of the iterate from the in-span stationary point at time $t\geq 1$. 
We further define:

\vspace{0.1cm}
\begin{definition}
	For a data matrix $\mathbf{X}$ and an adaptive method, with preconditioning matrices $\{\mathbf{D}(t): t\geq 1\}$, we call the adaptive method $(\alpha,\beta)$-converging on data, for any $\alpha, \beta \geq 0$, if and only if:
	$i)$ the out-of-span component of the pre-condition matrices decays as $|\lambda_{\max}(\tilde{\bs{D}}_2(t))| = \mathcal{O}\left(\dfrac{1}{t^\alpha}\right)$;
	$ii)$ the in in-span component of the iterates converges as, $e_{(1)}(t) = \mathcal{O}\left(\frac{1}{t^\beta}\right)$ (under  Eq.~\eqref{eq:linUpdate}).
\end{definition}

Any adaptive method with a pre-conditioner matrix that lies entirely in the span of the matrix will have $\alpha$ set to $\infty$. Full-matrix Adagrad, GD, and Newton all fall under this class of adaptive methods. 
Popular adaptive methods, such as diagonalized Adagrad, RMSProp, and methods with a diagonal pre-conditioner matrix with non-zero entries, the convergence depends on the rate of decay of both the $\tilde{\bs{D}}_2(t)$ as well as the rate of decay of the error of the in-span component.

\begin{theorem}\label{thm:unregularized}
	Consider the problem of over-parameterized linear regression with data matrix $\bs{X}$ and noise $\bs{\zeta}$ in the absence of regularization $\lambda = 0$. If the preconditioner matrix $\bs{D}(t)\succ 0$  $\forall t\geq 0$, and $\eta \in \left(0, \dfrac{2}{\lambda_{\max}(\bs{D}(t))\lambda_{\max}^2 (\bs{X})} \right),$ then in-span component of $\tilde{\bs{w}}(t)$ converges as follows
	\begin{align*}
	&\tilde{\bs{w}}_{(1)}(\infty) =
	(\tilde{\bsb}_{(1)}^* +\bs{\Lambda}_{(1)}^{-1} \bs{\zeta}_{(1)}).
	\end{align*} 
	Furthermore, for an adaptive method (in Eq.~\eqref{eq:linUpdate}) which is $(\alpha,\beta)$-converging on data,
	if $\alpha + \beta > 1$ the out-of-span component converges to a stationary point that satisfies
	$$
	\|\tilde{\bs{w}}_{(2)}(\infty) - \tilde{\bs{w}}_{(2)}(0)\|_2 \leq \mathcal{O}\left(\|\tilde{\bs{w}}_{(1)}(0)\|_2 + \frac{1}{\alpha+\beta-1}\right).
	$$
\end{theorem}
\textbf{Remark on Theorem~\ref{thm:unregularized}:} Let us deconstruct the claims made in Theorem \ref{thm:unregularized}. Theorem 1 says that if $\eta$ is set appropriately, then adaptive methods will perfectly fit the noisy training data. This is consistent with the claims in \citep{muthukumar2020harmless}. The convergence of out-of-span component depends on the decay rate of the pre-conditioner matrix $\tilde{\bs{D}}_2(t)$ as well as the decay rate of the error term in the in-span component $e_{(1)}(t) = \|(\bsb^*_{(1)} + \bs{\Lambda}^{-1}_{(1)} \bs{\zeta}_{(1)})  - \tilde{\bs{w}}_{(1)}(i)\|_2$. For the simple case, when $\beta\geq 1$ and $\tilde{\bs{D}}_{(2)}(t) = \bs{0}$ for all $t$, we have that the out-of-span component converges to  $\tilde{\bs{w}}_{(2)}(0)$. Next, if $\lim_{T\to \infty}\sum_{t=0}^{T} \max |\lambda(\tilde{\bs{D}}_{(2)}(t))| < \infty$ and $\alpha + \beta \geq 1$, then the out-of-span component converge. For all other cases, it is difficult to comment whether the out-of-span component will converge or diverge. Specifically, for $\alpha + \beta < 1$, we may not have divergence as the pre-conditioner matrices may align cancel the cumulative errors. 
\begin{figure}[!ht]
	\centering
	\includegraphics[scale=0.3]{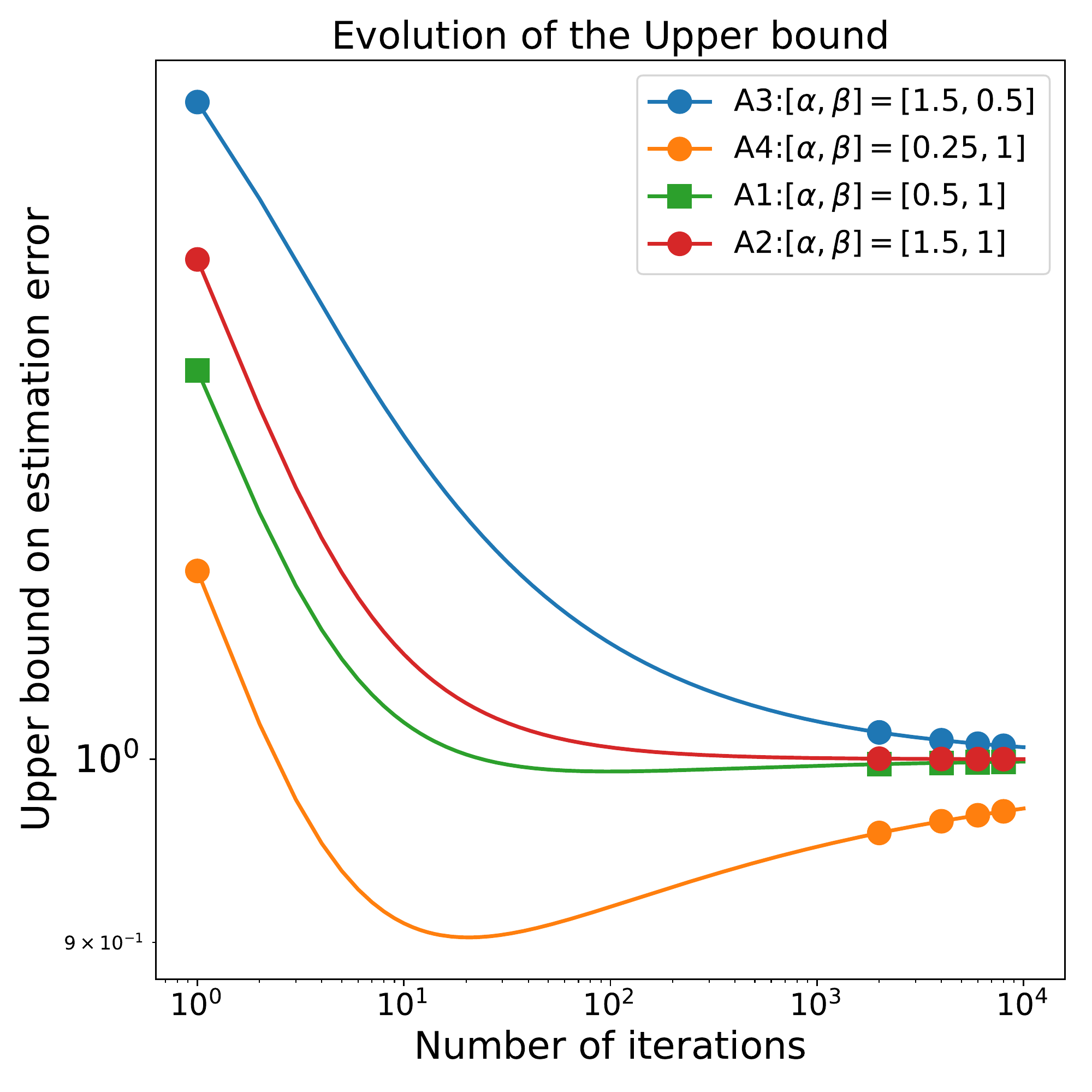}\\
	\caption{Evolution of upper bound dynamics for adaptive methods with different rates of $(\alpha, \beta)$ convergence, 
		with $a = 1$, $b = 0.7$, and $c = 0.1$.}
	\label{fig:synthetic}
\end{figure}

\textbf{Remark on Dynamics:} It is also interesting to note that the error in the in-span component converges to $0$ and the error in the out-of-span component increases and saturates to a particular value. Our derived upper bound on the training error at time $T$ for an adaptive method that is $(\alpha,\beta)$ converging on data is given as :
\begin{align*}
a + \dfrac{b}{(T+1)^{\beta}}\left(1 - \dfrac{c}{\alpha +\beta -1} \dfrac{1}{(T+1)^{\alpha -1}}\right)  
\end{align*}
for appropriate constants $a,b,c >0$. The dynamics is shown in the Figure~\ref{fig:synthetic}. 
Depending on the values of the constants, $\alpha$, and $\beta$, adaptive methods demonstrate variable convergence dynamics.
\section{Experiments}\label{sec:experiments} 
\blfootnote{Some of the experiments in this paper were also present in an earlier paper by us (Shah, V., Kyrillidis, A., Sanghavi, S. (2018). Minimum weight norm models do not always generalize well for over-parameterized problems. arXiv preprint arXiv:1811.07055) which had a different flavor of theoretical results. The experiments have been reused as they reinforce some of the claims made in this paper}

In this section, we focus on replicating the theoretical claims made in the previous parts using synthetic experiments for over-parameterized linear regression. Next, we show that these observations can be extended to the deep learning setup as well.
We empirically compare two classes of algorithms:
\begin{itemize}[leftmargin=0.5cm]
	\item  Plain gradient descent algorithms, including the mini-batch stochastic gradient descent and the accelerated stochastic gradient descent, with constant momentum.
	\item Adaptive methods like AdaGrad \citep{duchi2011adaptive}, RMSProp \citep{tieleman2012lecture}, and Adam \citep{kingma2014adam}, and the AdaGrad variant. Adaptive methods can include anything with a time-varying pre-conditioner matrix. 
\end{itemize}

\subsection{Linear Regression} 

In the first part, we consider a simple linear regression example generated with where the elements of both $\bs{X}$ and $\bs{\zeta}$ are generated using from $\mathcal{N}(0,1)$ distribution in an i.i.d. manner. The test data is sampled from the same distribution as training data. Here, we show that different adaptive methods can yield better performance in terms of  generalization.
The expressions for AM1, AM2 and AM3 are given as follows:
\begin{figure*}[!ht]
	\centering
	\includegraphics[scale=0.23]{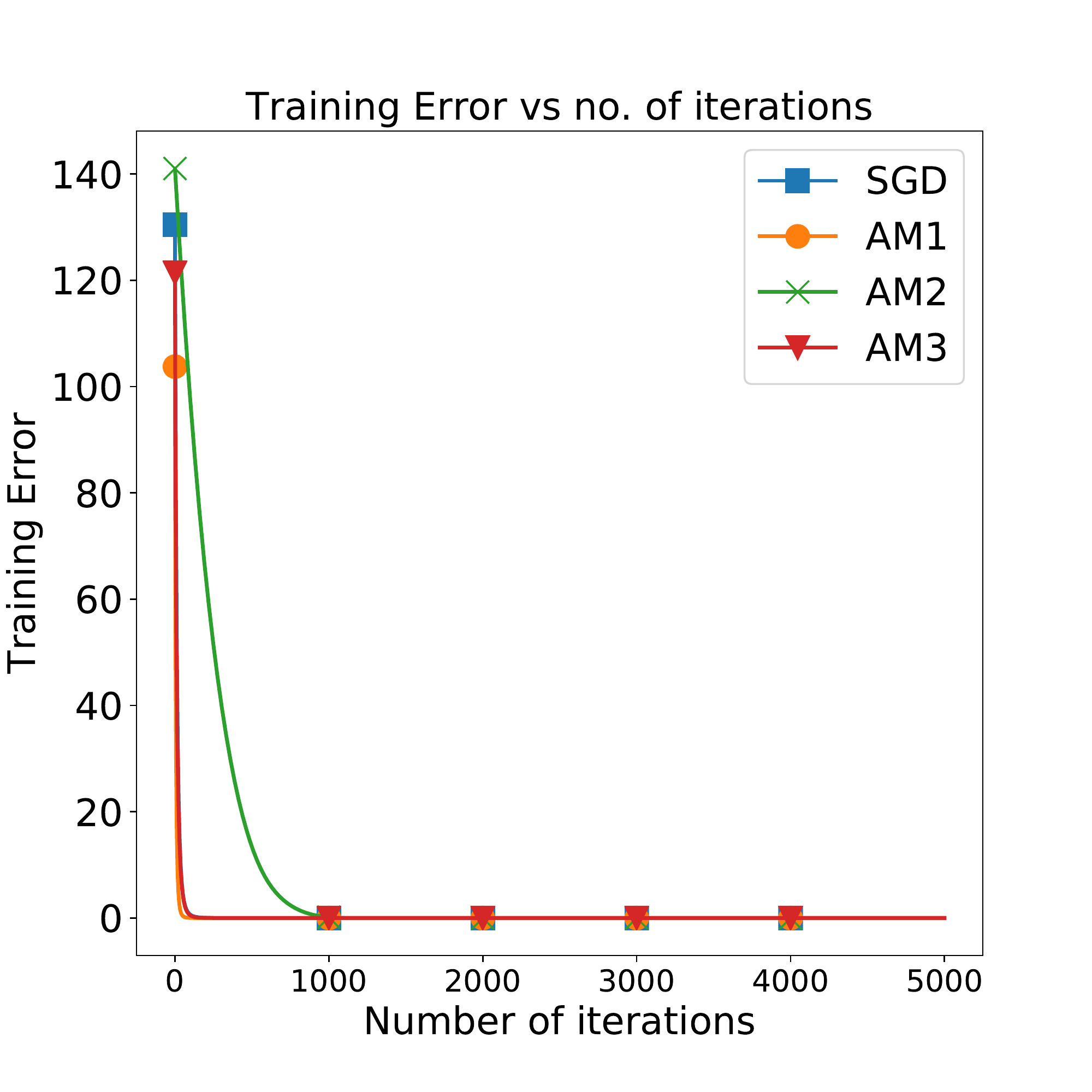} 
	\includegraphics[scale=0.23]{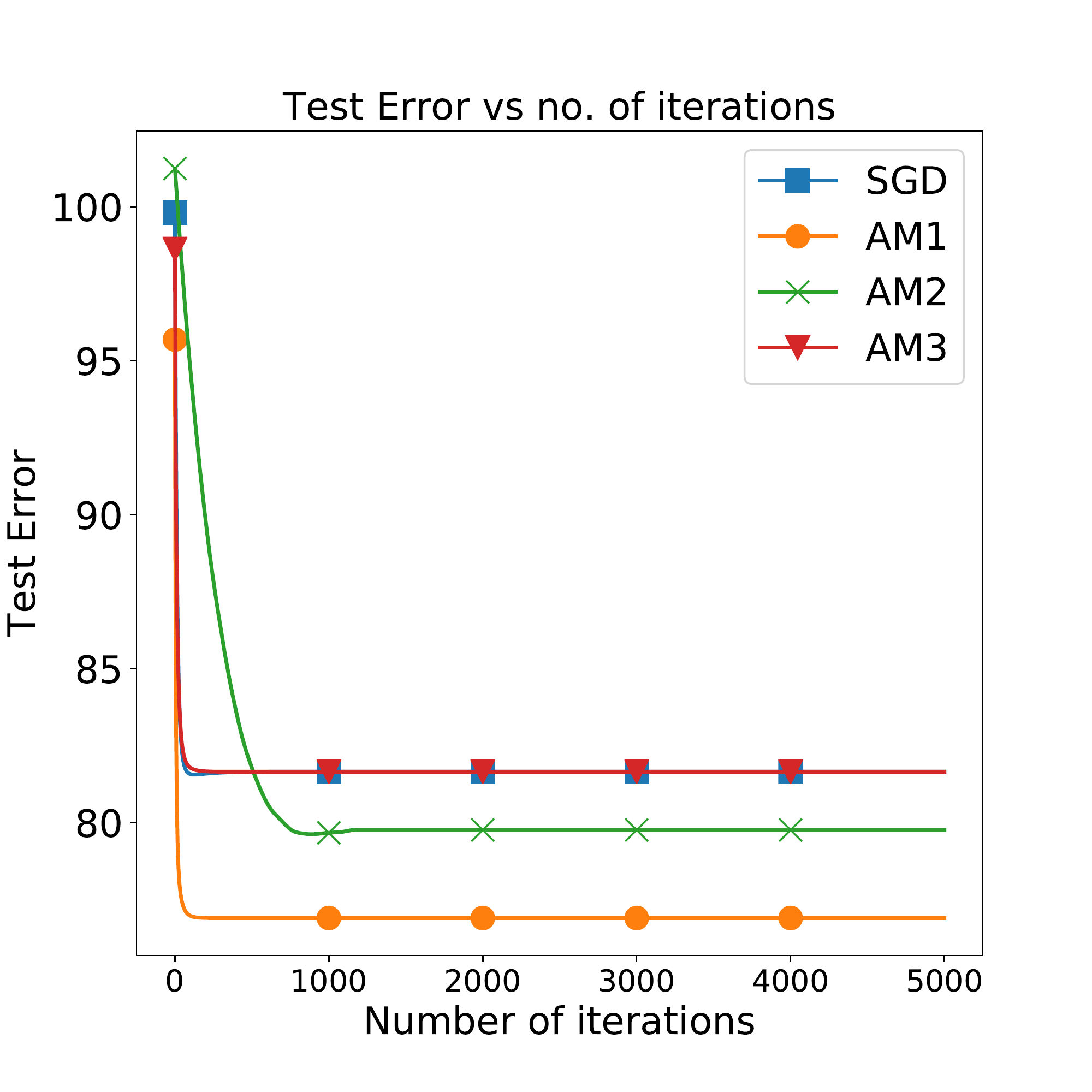} 
	\includegraphics[scale=0.23]{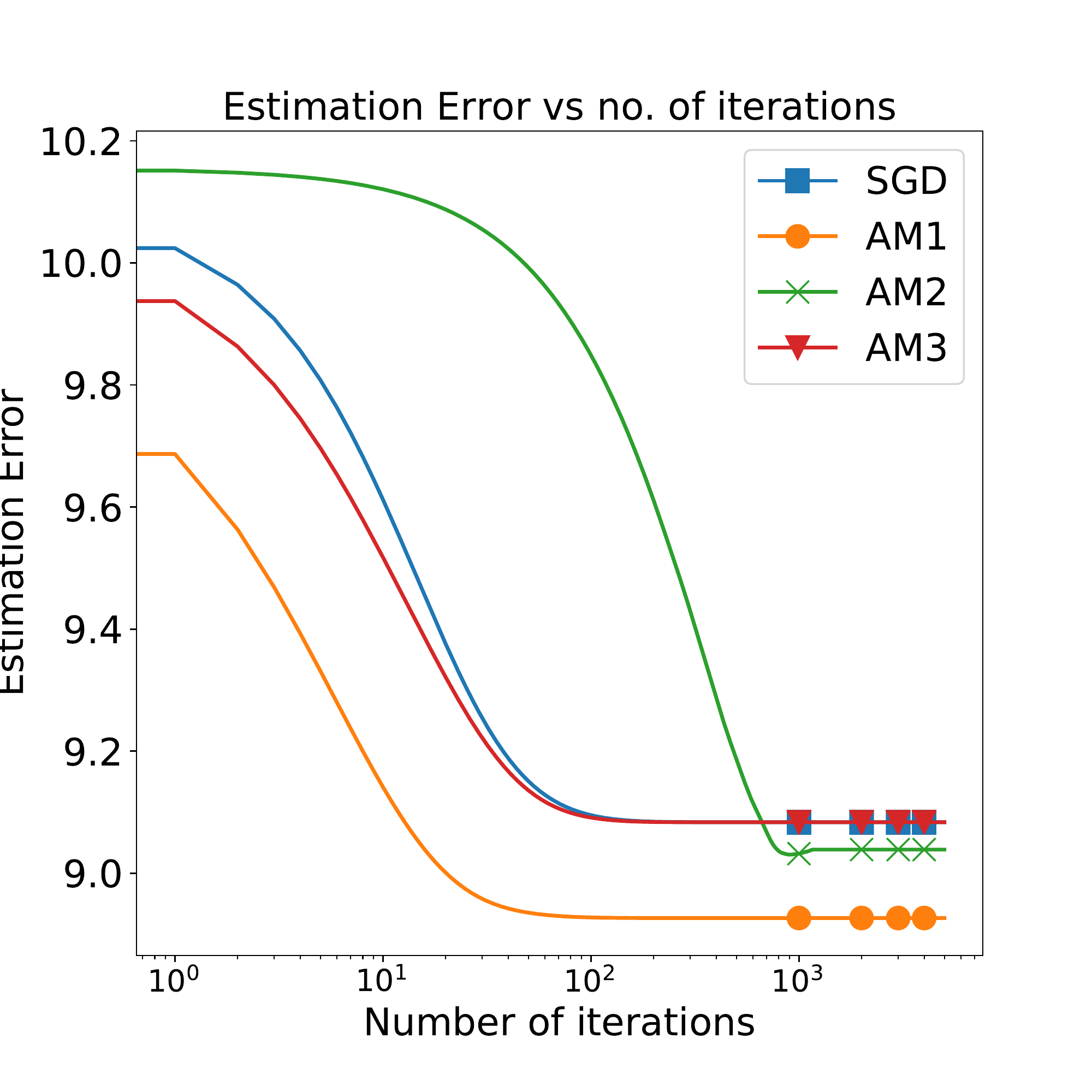}
	\caption{Synthetic example of over-parameterized linear regression where adaptive methods show better test error performance. Notice that adaptive method AM1 not only allows us to achieve faster convergence but also better generalization. Estimation error, $\|\|\bs{w}(t) - \bsb^*\|\|$ is in the semilog scale on the x axis (to highlight the double descent like phenomena in AM2 as predicted by the Remark at the end of Section 2). The reported results are the average of 5 runs with different initializations for a given realization of data.}
	\label{fig:lin_example}
\end{figure*}

\begin{table*}[!t]
	\centering
	\vspace{0.3cm}
	\begin{footnotesize}
		\begin{tabular}{c c c c c c c}
			\toprule 
			$d=50$  & \phantom{5} & GD &  AM1 & AM2  &AM3\\  \midrule
			$\bs{D}(t)$&   &$\bs{I}$ & $\bar{\bs{D}}_1(t)$ &$\bs{D}_2(t)$ &$\bar{\bs{D}}_3(t)= \mathcal{P}_{\bs{X}}(\bar{\bs{D}_1}(t))$
			\\
			\midrule
			\multirow{3}{*}{$n = 10$} & Training Error &$1.27\cdot 10^{-28}$ &$\mathbf{1.42\cdot 10^{-29}}$ &$8.64\cdot 10^{-4}$ &$8.64\cdot 10^{-29}$ \\
			\phantom{1} & Test Error  &$81.56$ &$\mathbf{76.94}$ &$79.62$ &$81.65$\\
			\phantom{1} & $\|\bs{w} - \bs{w}^*\|$ &$9.08$ &$\mathbf{8.92}$ &$9.03$ &$9.08$\\
			\midrule
			\multirow{3}{*}{$n = 40$} & Training Error &$4.77\cdot 10^{-5}$ &$\mathbf{6.07\cdot 10^{-7}}$ &$3.31\cdot 10^{-3}$ &$8.64\cdot 1.17^{-4}$ \\
			\phantom{1} & Test Error &$\mathbf{18.62}$ &$19.56$  &$20.35$ &$18.65$ \\
			\phantom{1} & $\|\bs{w} - \bs{w}^*\|$ &$\mathbf{4.31}$ &$4.37$ &$4.51$ &$\mathbf{4.31}$\\
			\bottomrule
		\end{tabular}
	\end{footnotesize}\caption{Illustrating the varying performances of adaptive methods for over-parameterized linear regression. The final values are the average of 5 runs. AM1: Diagonalized Adagrad, AM2: Adagrad (AM1) Variant (where we square the diagonal terms instead of taking the square root), AM3: Projected version of AM1 onto the span of X. For AM3, $\bs{\tilde{D}}_2(t) = 0,~ \forall~ t$ and consistent with Theorem $\ref{thm:unregularized}$ it converges to the same point as SGD. AM1 and AM2 satisfy the $(\alpha, \beta)$ convergence criterion leading to convergence to a different point and different  generalization than SGD.} \label{table:mysterysolved}
\end{table*}

\begin{table*}[!t] 
	\centering
	\caption{Prediction accuracy and distances from the minimum norm solution for plain gradient descent and adaptive gradient descent methods. Adagrad variant squares the pre-conditioner matrix values instead of taking the square root in Adagrad. The distances shown are median values out of 100 different realizations for each setting; the accuracies are obtained by testing $10^4$ predictions on unseen data.} \label{table:counterex11}
	\vspace{0.3cm}
	\begin{footnotesize}
		\begin{tabular}{c c c c c c c c c c c}
			\toprule
			\phantom{1} & & \phantom{3} & & \phantom{5} & & Gradient Descent & &  AdaGrad variant & & Adam \\
			\midrule
			\multirow{8}{*}{$n = 10$} & & \multirow{2}{*}{$\ell = 1/32$} & &Test Acc. (\%) & &63 & &\textbf{100}  & &91   \\ 
			\phantom{1}		& & 		\phantom{3}		& & $\|\widehat{\bs{w}} - \bsb^*\|_2$ & &$1.015\cdot 10^{-16}$ & &$4.6924 \cdot 10^4$  & &$0.1007$   \\
			\cmidrule{7-11}
			& & \multirow{2}{*}{$\ell = 1/16$} & &Test Acc. (\%)  & &53 & &\textbf{100}  & &87 \\ 
			& & 						& &  $\|\widehat{\bs{w}} - \bsb^*\|_2$ & &$1.7401\cdot 10^{-16}$ & &$1.1504\cdot 10^3$   & &$0.0864$  \\
			\cmidrule{7-11}
			& & \multirow{2}{*}{$\ell = 1/8$} & &Test Acc. (\%)  & &58 & &\textbf{99}  & &84 \\ 
			& & 						& &  $\|\widehat{\bs{w}} - \bsb^*\|_2$ & &$4.08\cdot 10^{-16}$ & &$112.03$  & &$0.0764$ \\
			\midrule
			\multirow{8}{*}{$n = 50$} & & \multirow{2}{*}{$\ell = 1/32$} & & Test Acc. (\%) & &77 & &\textbf{100} & &88 \\
			\phantom{1}		& & 		\phantom{3}		& &  $\|\widehat{\bs{w}} - \bsb^*\|_2$ & &$4.729\cdot 10^{-15}$ & &$3.574\cdot 10^3$   & &$0.0271$\\
			\cmidrule{7-11}
			& & \multirow{2}{*}{$\ell = 1/16$} & & Test Acc. (\%)& &80 & &\textbf{100}  & &89\\ 
			& & 						& &  $\|\widehat{\bs{w}} - \bsb^*\|_2$& &$6.9197\cdot 10^{-15}$ & &$4.44\cdot 10^2$   & &$0.06281$\\
			\cmidrule{7-11}
			& & \multirow{2}{*}{$\ell = 1/8$} & &Test Acc. (\%) & &91 & &\textbf{100}   & &89 \\
			& & 						& &  $\|\widehat{\bs{w}} - \bsb^*\|_2$ & &$9.7170\cdot 10^{-15}$ & &$54.93$   & &$0.1767$ \\
			\bottomrule
		\end{tabular}
	\end{footnotesize}
\end{table*}

\begin{small}
	\begin{align*}
	\bs{D}_{1}(t) &= \texttt{diag}\left( \dfrac{1}{\sum_{j = t - J}^t \nabla f(\bs{w}(j)) \odot\nabla f(\bs{w}(j)) + \varepsilon}\right) \notag\\
	\bs{D}_{2}(t)	&=\texttt{diag}\left(\dfrac{1}{\sum_{j = t - J}^t \nabla f(\bs{w}(j)) \odot\nabla f(\bs{w}(j)) + \varepsilon}\right) \notag\\
	\bs{D}_{3}(t) &= \mathcal{P}_{\bs{X}}(\bs{D}_1(t))
	\end{align*}
\end{small}
Here, we assume that $	\bs{D}_{i}(t) \succ 0, \text{for some} ~\varepsilon > 0,	~~\text{and}~~ J < t \in \mathbb{N}_{+} \quad \forall~i$.
In Figure \ref{fig:synthetic}, we observe that the conditions described in previous section allow us to predict the performance of adaptive methods relative to gradient descent based methods, i.e. we can ascertain whether a given adaptive method will have the same generalization performance as SGD based methods or not. 
\paragraph{Classification:}
In this example, we consider a linear regression problem with binary outputs i.e. $y_i \in \{-1, 1\}$ (a variant of the example proposed in \citep{wilson2017marginal}).
\begin{align} 
\left(x_i\right)_j &= 
\begin{cases}
y_i \ell, & \!\!j = 1, \\
1, & \!\!j = 2, 3,\\
1, &  \!\!j=4+ 5(i-1), \\
0, & \!\!\text{otherwise}.
\end{cases}  ~~~~~ \text{if} ~ y_i = 1, 
\\
\left(x_i\right)_j &= 
\begin{cases}
y_i\ell, & \!\!j = 1, \\
1, & \!\!j = 2, 3, \\
1, & \!\!j=  4+ 5(i-1),\\
& \hspace{0.3cm}  \cdots, 8+ 5(i-1), \\
0, & \text{otherwise}.
\end{cases} ~~~ \text{if} ~ y_i = -1. \label{eq:features}
\end{align}
The expressions for the pre-conditioner matrix of Adagrad variant is:
\begin{small}
	\begin{align*}
	\bs{D}_{AV}(t) &= \texttt{diag}\left(\dfrac{1}{\sum_{j = t - J}^t \left(\nabla f(\bs{w}(j)) \odot\nabla f(\bs{w}(j))\right)^2 + \varepsilon}\right) 
	\end{align*}
\end{small}
Note that $	\bs{D}_{AV}(t) \succ 0, \text{for some} ~\varepsilon > 0,	~~\text{and}~~ J < t \in \mathbb{N}_{+}$.  Table \ref{table:counterex11} depicts that even in terms of test accuracy, adaptive algorithms can yield better generalization performance than SGD. This supports the claim made recently in \citep{muthukumar2020classification} for non-adaptive methods; \textit{the testing criterion can play a crucial role in determining the generalization performance}. We observe that this claim holds for adaptive methods as well.

\begin{figure*}[!htbp]    
	\centering
	\includegraphics[width=0.23\textwidth]{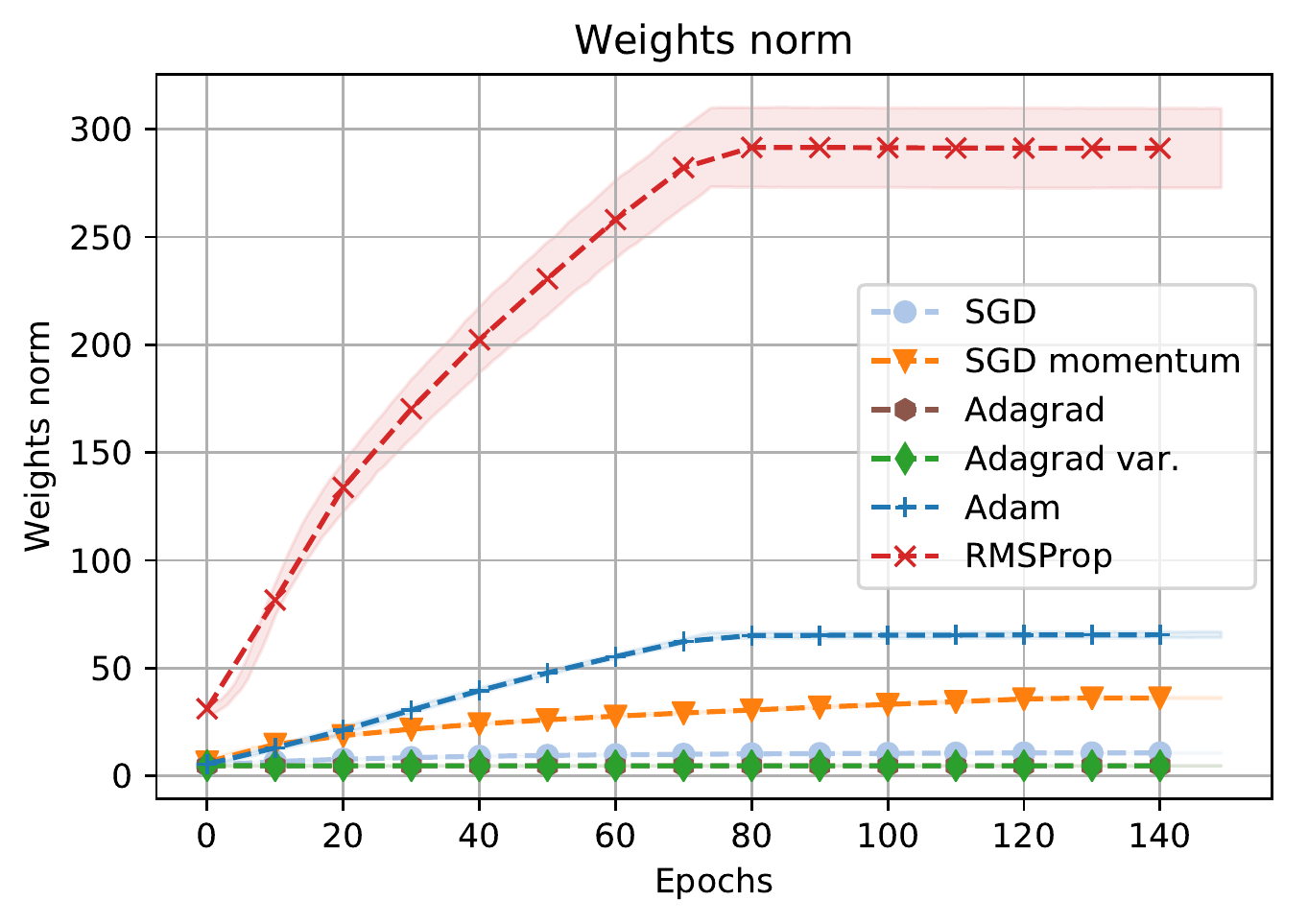}
	\includegraphics[width=0.23\textwidth]{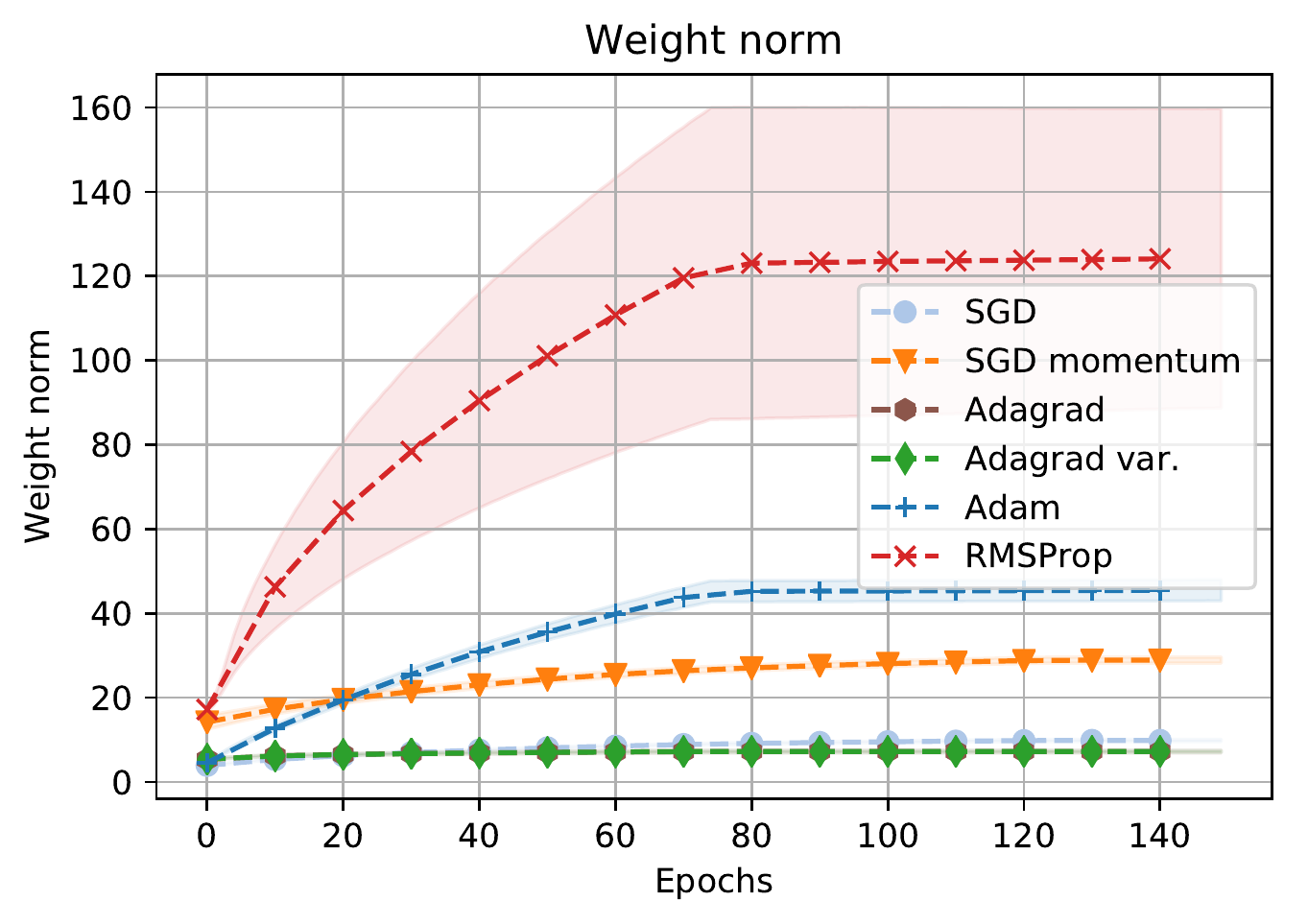}
	\includegraphics[width=0.23\textwidth]{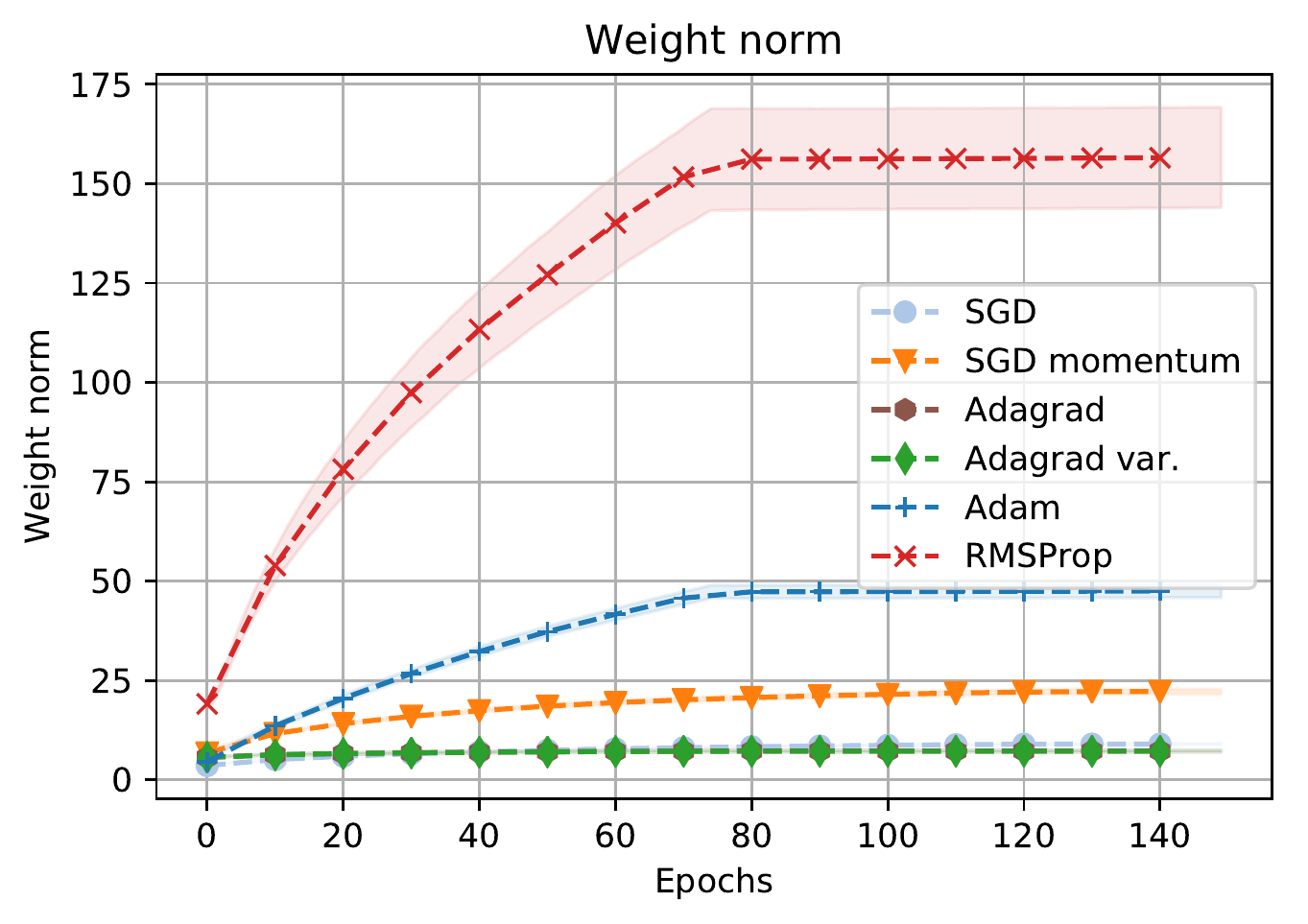}
	\includegraphics[width=0.23\textwidth]{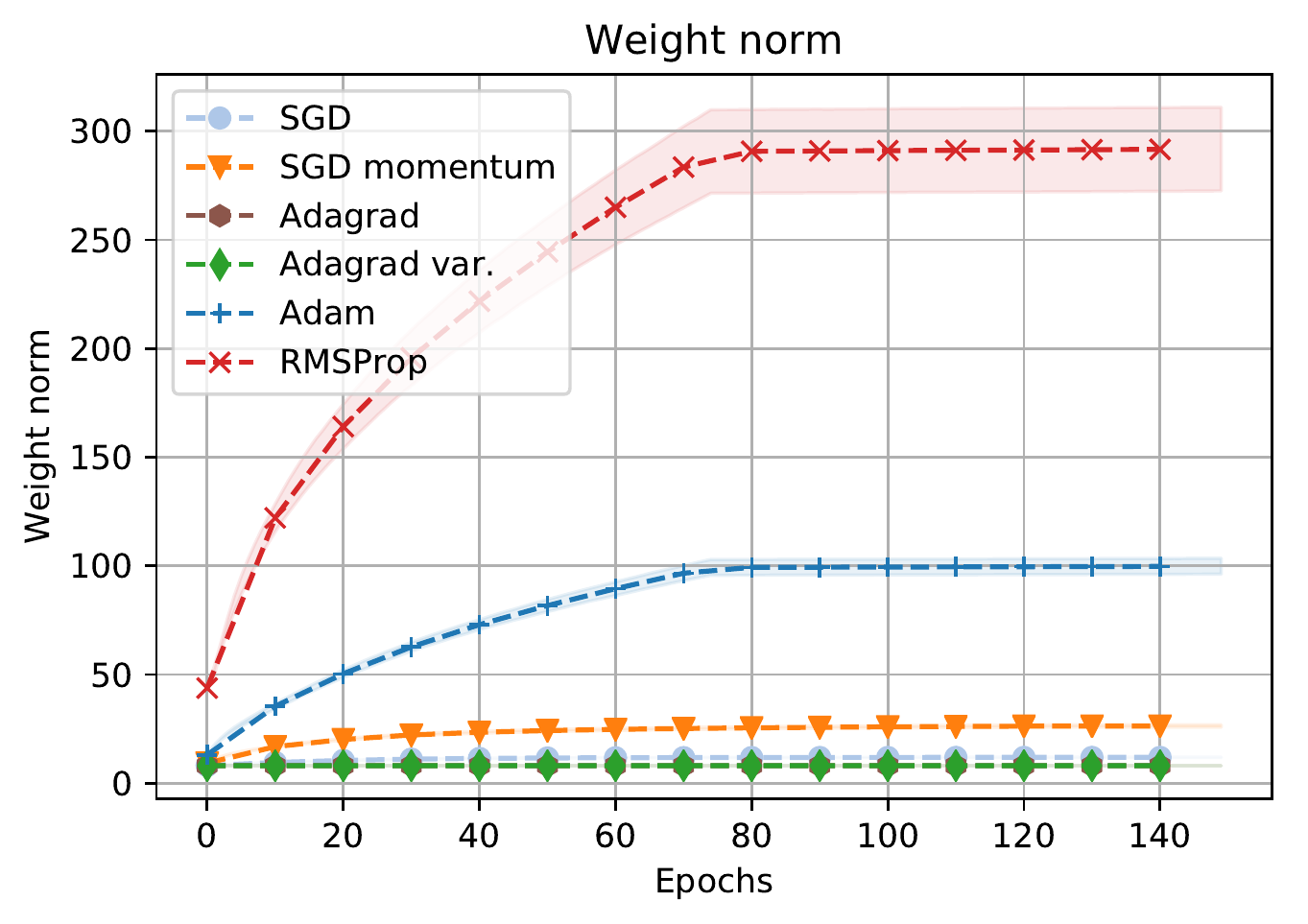}
	\\
	\includegraphics[width=0.23\textwidth]{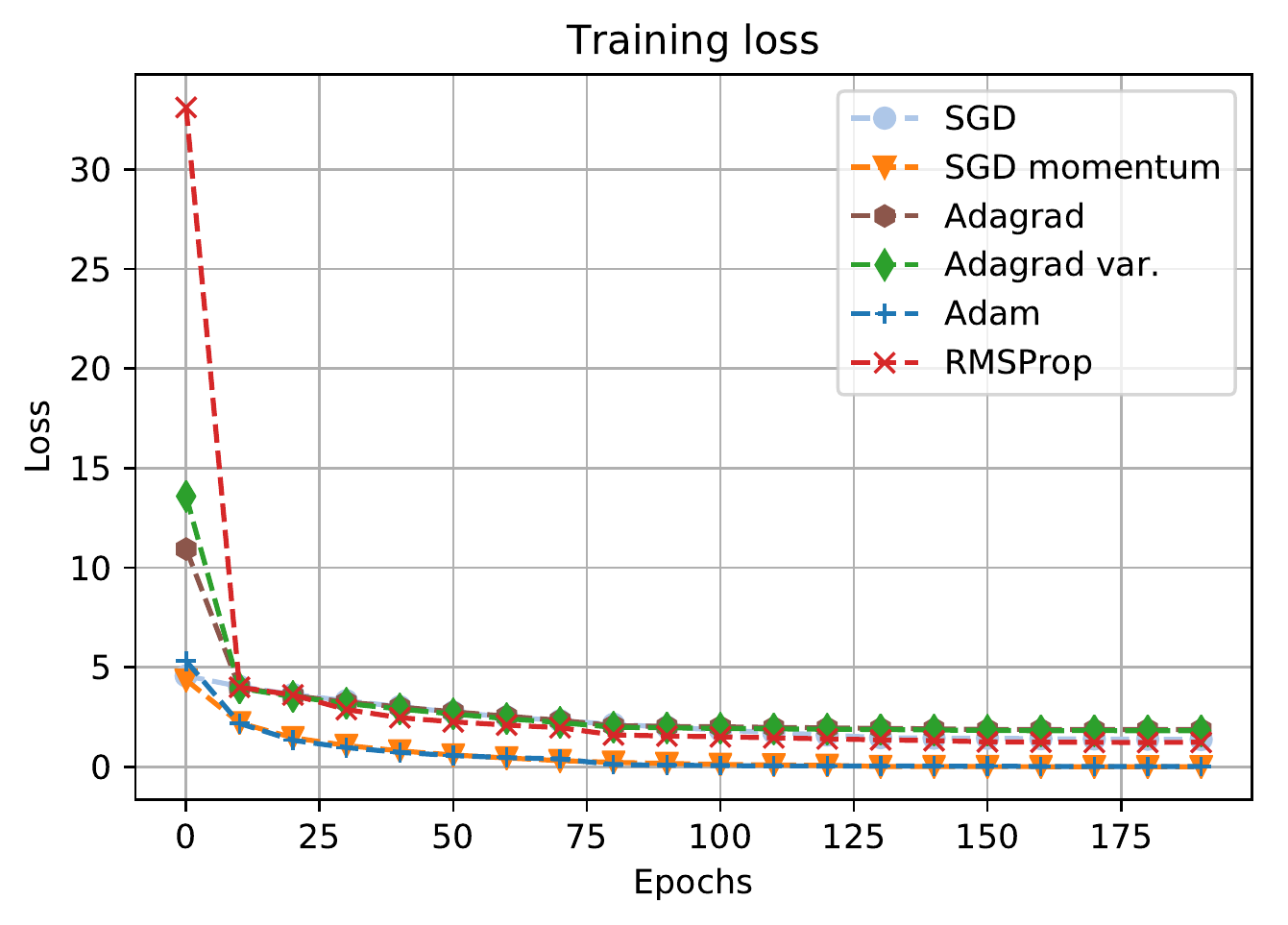}
	\includegraphics[width=0.23\textwidth]{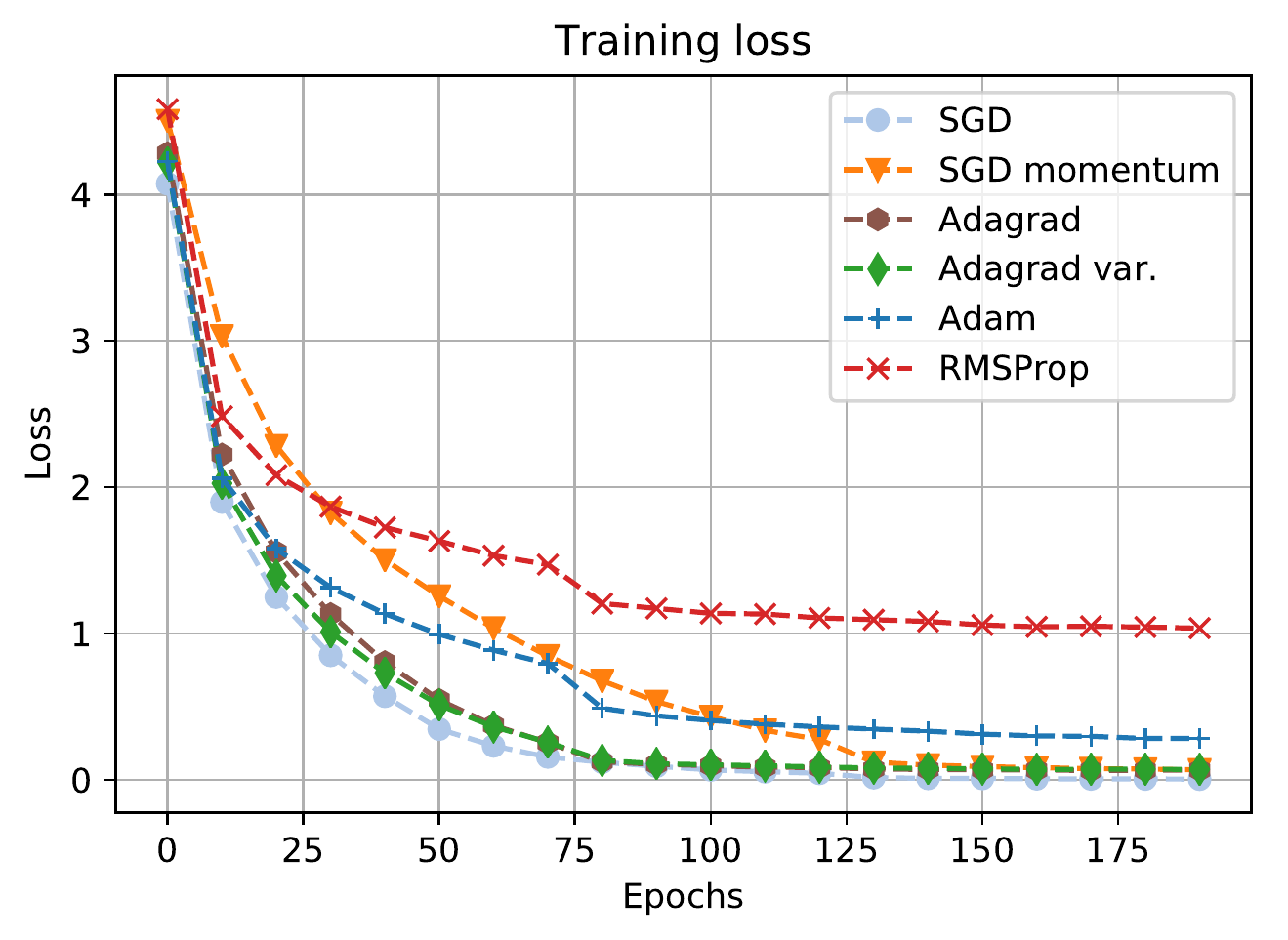}
	\includegraphics[width=0.23\textwidth]{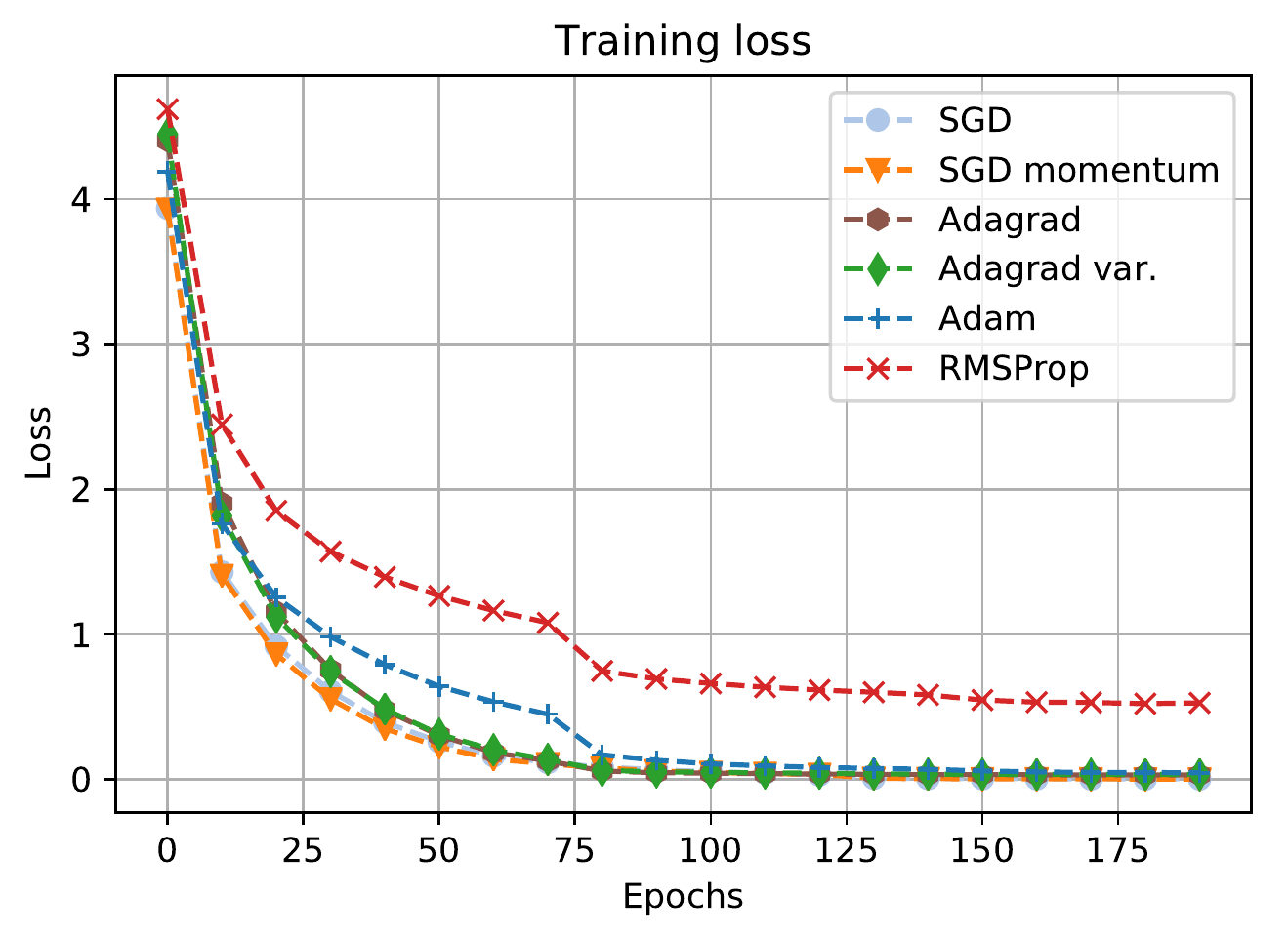}
	\includegraphics[width=0.23\textwidth]{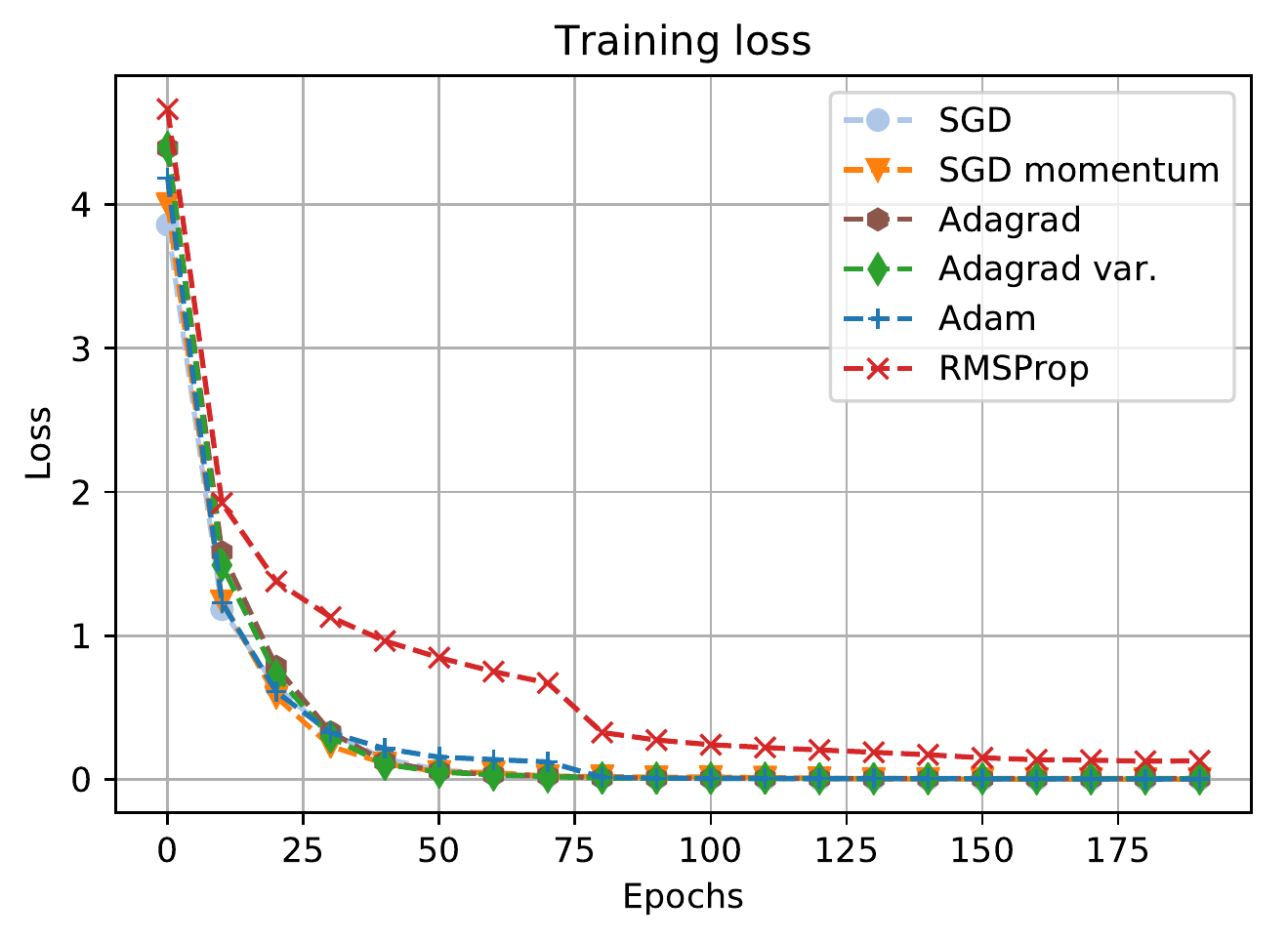}
	\\
	\includegraphics[width=0.23\textwidth]{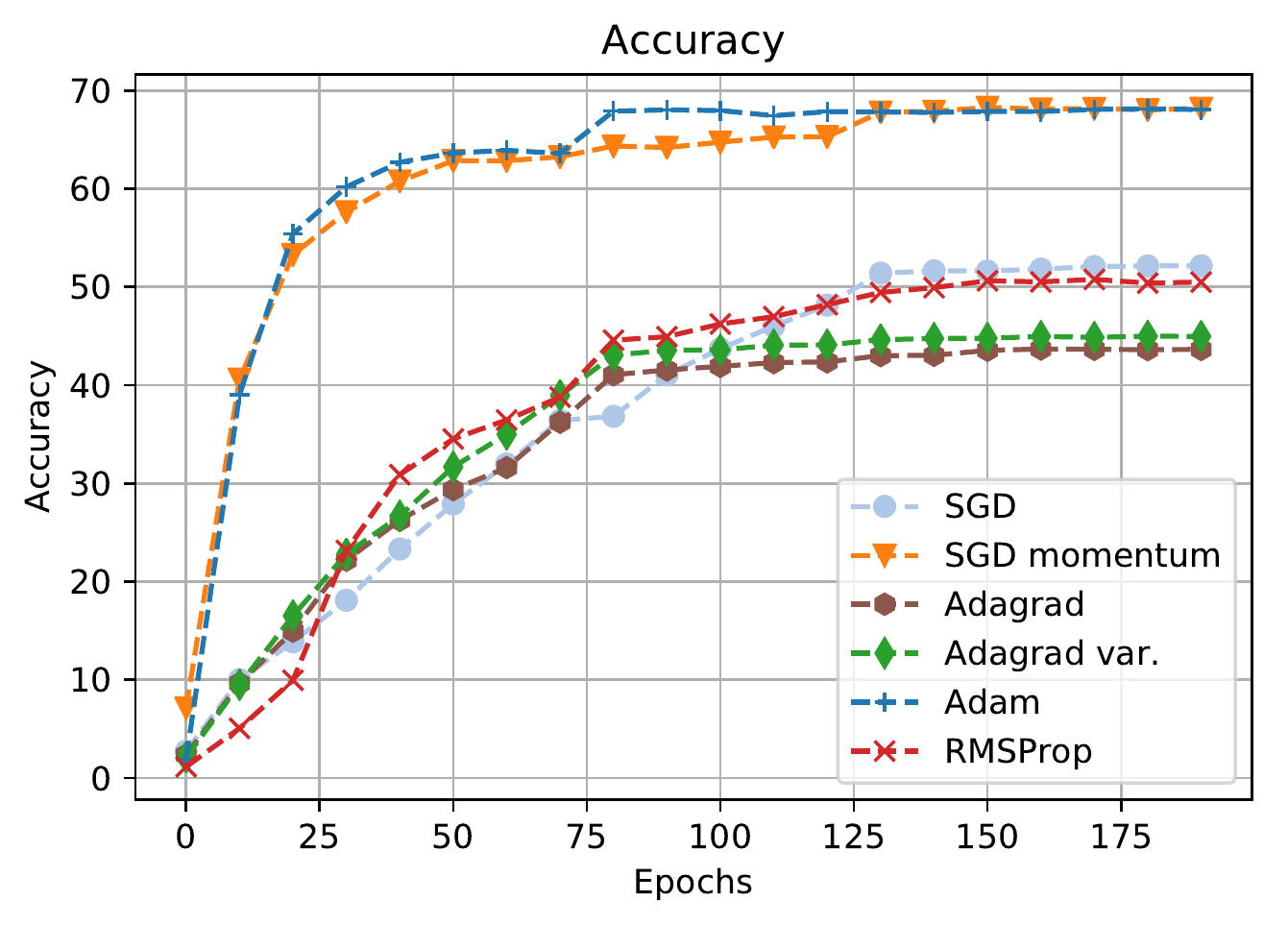} 
	\includegraphics[width=0.23\textwidth]{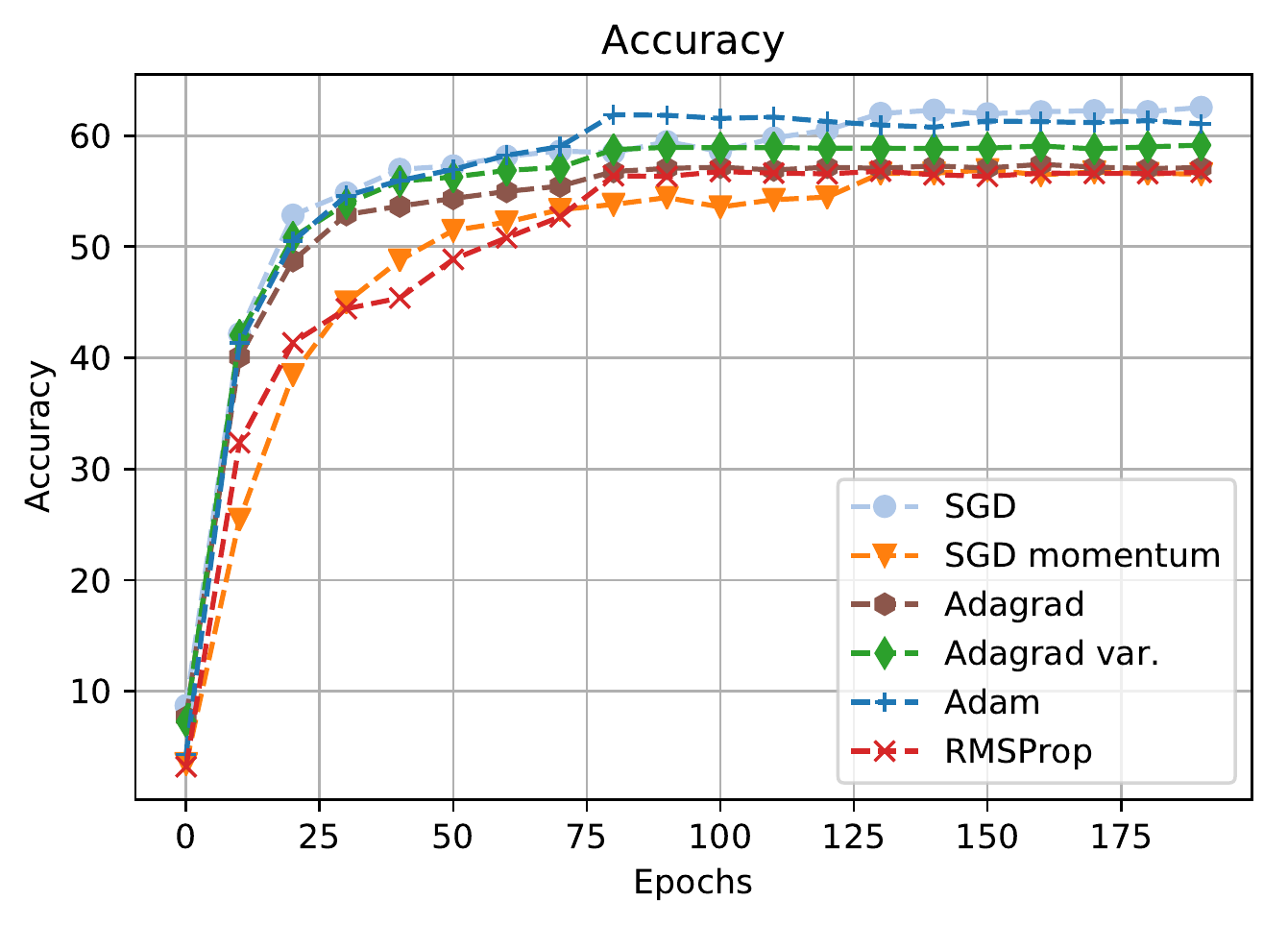}    
	\includegraphics[width=0.23\textwidth]{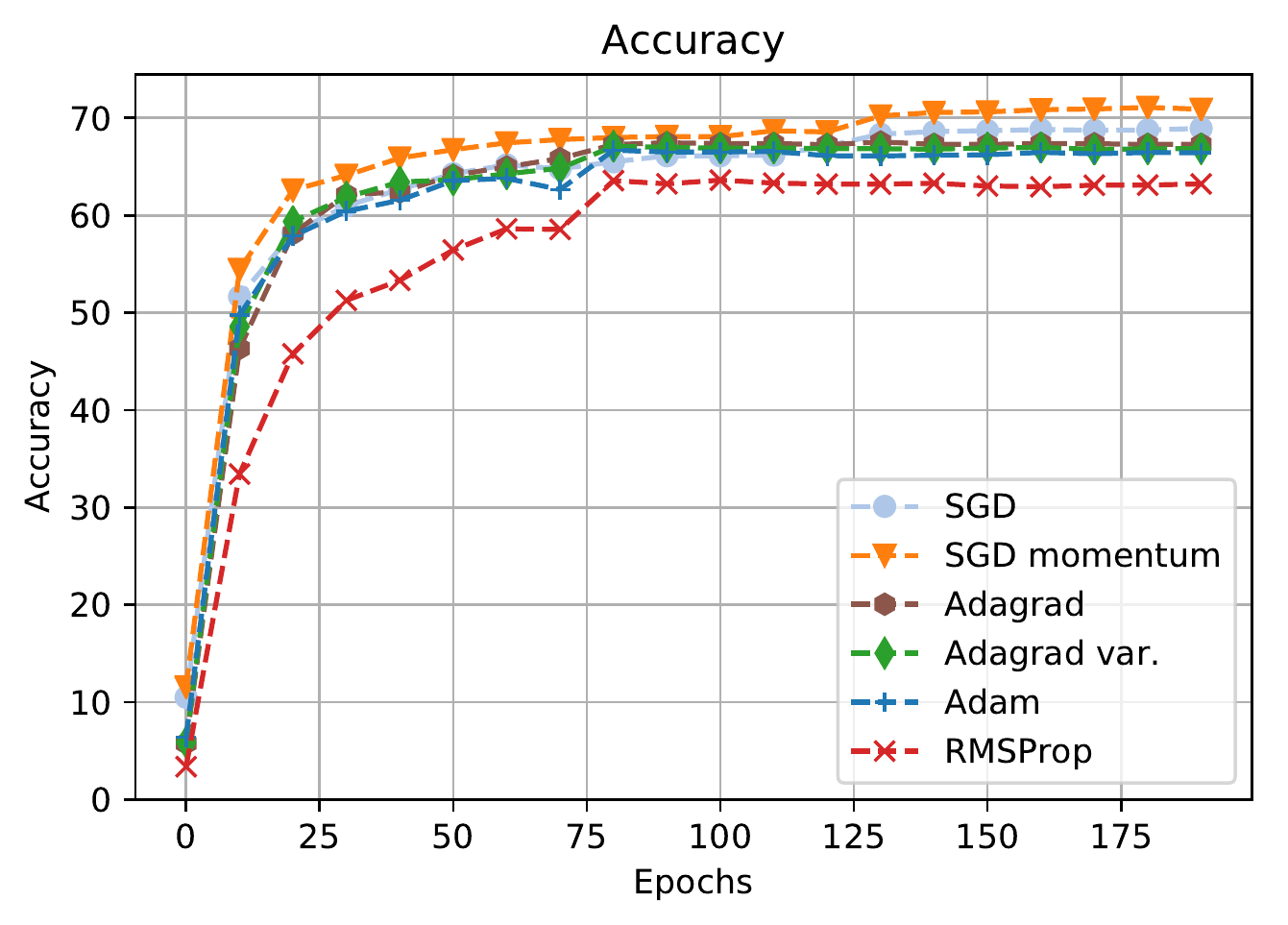}    
	\includegraphics[width=0.23\textwidth]{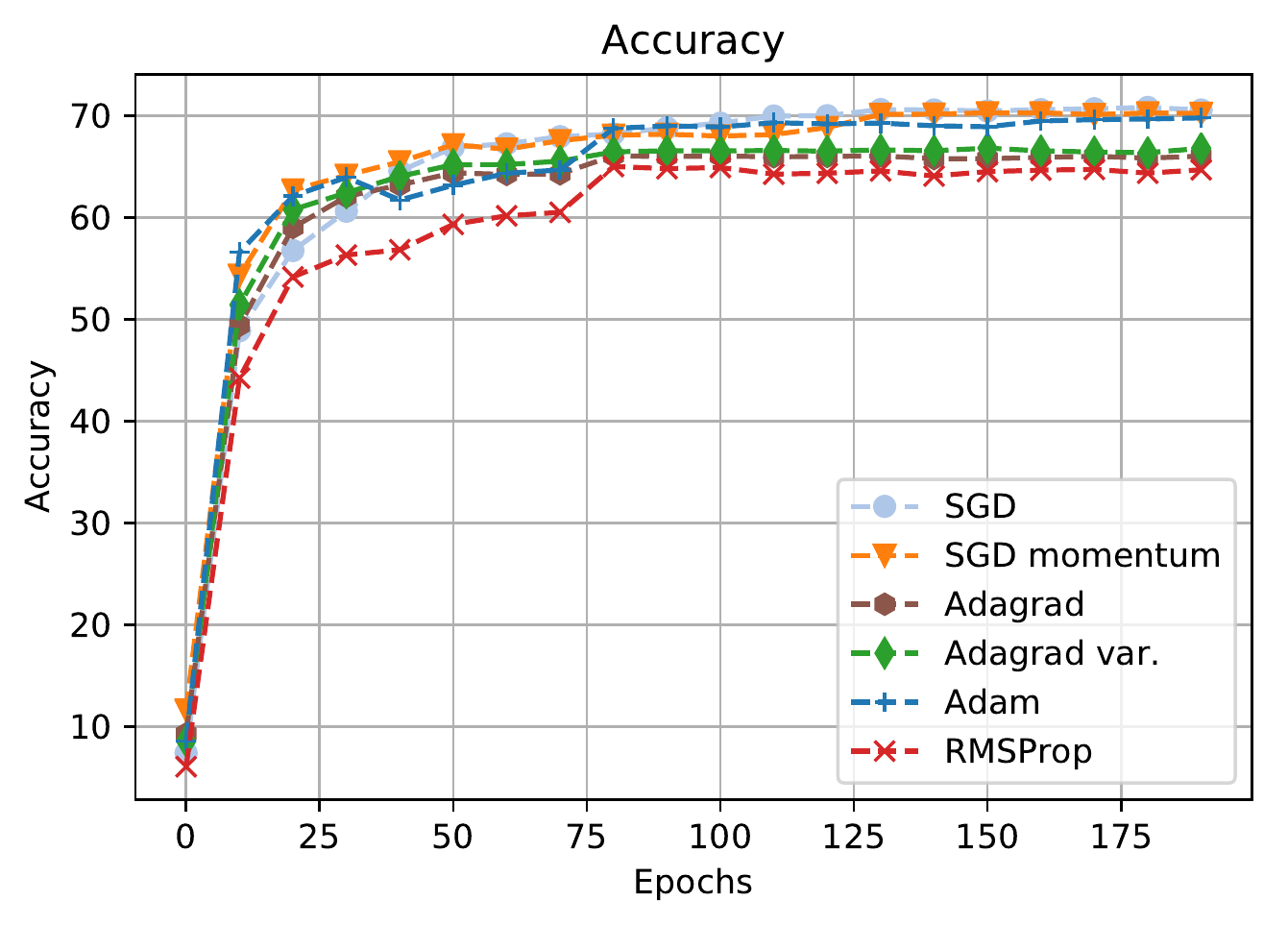}    
	\caption{Accuracy results on unseen data, for different NN architectures on CIFAR100. \emph{Left panel:} Accuracy and training loss for PreActResNet18 in \cite{he2016identityb}; \emph{Left middle panel:} Accuracy and training loss for MobileNet in \cite{howard2017mobilenets}; \emph{Right middle panel:} Accuracy and training loss for MobileNetV2 in \cite{sandler2018inverted}, \emph{Right panel:} Accuracy and training loss for GoogleNet in \cite{sandler2018inverted}. \emph{Top row:} Weight vectors of the last layer, \emph{Middle row:} Training Loss, \emph{Last row:} Test Accuracy.} \label{fig:01}
\end{figure*}

\subsection{Deep Learning}
In this experiment, we observe that the theoretical claims made for the generalization of adaptive methods for over-parameterized linear regression extend over to over-parameterized neural networks. We perform extensive experiments on CIFAR-100 in Figure \ref{fig:01}. For CIFAR-100 datasets we explore four different architectures;  PreActResNet18 \cite{he2016identityb}, MobileNet \cite{howard2017mobilenets}, MobileNetV2 \cite{sandler2018inverted}, GoogleNet \cite{sandler2018inverted}.
After a hyper-parameter tuning phase, we selected the best choices among the parameters tested.
The results show no clear winner once again, which overall support our claims: \emph{the superiority depends on the problem/data at hand; also, all algorithms require fine tuning to achieve their best performance.}

An important observation of Figure \ref{fig:01} comes from the top row of the panel.
There, we plot the Euclidean norm $\|\cdot\|_2$ of all the trainable parameters of the corresponding neural network. 
While such a norm could be considered arbitrary (e.g., someone could argue other types of norms to make more sense, like the spectral norm of layer), we use the Euclidean norm as $\pmb{i)}$ it follows the narrative of algorithms in linear regression, where plain gradient descent algorithms choose minimum $\ell_2$-norm solutions, and $\pmb{ii)}$ there is recent work that purposely regularizes training algorithms towards minimum norm solutions \citep{bansal2018minnorm}. 

Our findings support our claims: in particular, for the case of MobileNet and MobileNetV2, Adam, an adaptive method, converges to a solution that has at least as good generalization as plain gradient methods, while having $2 \times$ larger $\ell_2$-norm weights.
However, this may not always be the trend: in Figure \ref{fig:01}, left panel, the plain gradient descent models for the PreActResNet18 architecture \citep{he2016identityb} show slightly better performance, while preserving low weight norm.
The same holds for  GoogleNet; see Figure \ref{fig:01}, right panel.

Another observation is that like linear regression, here also adaptive methods can be clubbed into different categories based on the evolution of their pre-conditioner matrices, $\bs{D}(t)$.  It is evident that algorithms that have similar weight norms have similar training loss performance; however the other side of the claim need not be true. The experiments indicate the potential of adaptive methods to show better generalization w.r.t. unseen data. 
The details about experiments on more datasets, hyperparameter tuning, practical issues in implementation are available in the Appendix.

\section{Conclusions}
In this paper, we consider two class of methods described: non-adaptive methods (Eq.  \eqref{eq:nonadaptive}) and adaptive methods (Eq. \eqref{eq:adaptive}). Switching to a spectral domain allows us to divide adaptive methods into two further categories based on if they will have the same generalization as SGD or not (assuming the same initialization point). We obtain that the convergence of adaptive methods completely depends on the structure of pre-conditioner matrices $\bs{D}(t)$ along with the initialization point and the given data.

Our theoretical analysis allows us to obtain useful insights into the convergence of adaptive methods, which can be useful while designing new adaptive methods. If the aim while designing an adaptive method is faster convergence and similar generalization as SGD, then it is important to ensure that the pre-conditioner matrix lies in the span of the data matrix $\bs{D}(t) = \mathcal{P}_{\bs{X}}(\cdot)$. Examples of such $\bs{D}(t)$ include $\{\mathbb{I}, (\bs{X}^\top \bs{X})^{-1}\}$. 
However, if the aim is to hope for a different generalization than SGD (if SGD gets stuck on specific bad minima), then it is essential to ensure that the conditions in Theorem \ref{thm:unregularized} are satisfied to ensure that $\tilde{\bs{w}}(t)$ converges to a different solution. Our experimental results on over-parameterized settings for both linear regression and deep learning back our theoretical claims.


\bibliographystyle{apalike}
\bibliography{references.bib}

\onecolumn

\appendix
\section{Supplementary material}

\subsection{Proof of Proposition~\ref{lem:spectralRegularized}}
From Theorem~\ref{thm:convseq},  we know that a sufficient condition for the convergence under update \eqref{eq:linUpdate} is 
$$\sup_{t\geq 1} |\lambda\left(I - \eta \tilde{\bs{D}}(t) (\bs{\Lambda}^2 + \lambda \bs{I})\right)| < 1.$$ 
Thus, for $\lambda > 0$, the dynamics converges to a bounded weight vector for any \\
$\eta \in \left(0,  2\left(\lambda_{max}(\bs{D}(t))(\lambda_{max}^2 (\bs{X})+ \lambda)\right)^{-1}\right).$

We now characterize the fixed point of the dynamics  in~\ref{eq:linUpdate}. When the convergence happens, 
for any fixed point $\hat{\bsb}$ of the updates  in~\ref{eq:linUpdate}  $$\bs{D}(t) \left( \lambda \hat{\bsb} +  \bs{X}^T\bs{X}(\hat{\bsb}- \bsb^*) - \bs{X}^T\bs{w} \right) = \bs{0}.$$ Because, $\inf_t rank(\bs{D}(t)) = d$ (full rank) we must have 
${\lambda \hat{\bsb} +  \bs{X}^T\bs{X}(\hat{\bsb}- \bsb^*) - \bs{X}^T\bs{w} = \bs{0}}$. Expanding the l.h.s. in terms of the SVD of the data matrix we obtain, 
\begin{align*}
\sum_{r=1}^{d} \lambda \tilde{\bs{w}}_r \bs{v_r} + \sum_{r=1}^{R} \left(\lambda_r^2 \tilde{\bs{w}}_r - \lambda_r^2\tilde{\bs{w}}^*_r  - \lambda_r \tilde{\bs{\zeta}}_r \right) \bs{v}_r = \bs{0}.
\end{align*}
Therefore, for $\lambda \geq 0$ (holds for both regularized and unregularized) we have $\bs{v_r}^T\hat{\bsb} = \tfrac{ \lambda_r^2 \tilde{\bs{w}}^*_r + \lambda_r \tilde{\bs{\zeta}}_r }{ \lambda+  \lambda_r^2 }$ for $r\leq R$. Further, for $\lambda > 0$, $\bs{v_r}^T\hat{\bsb} = 0$ for $r\geq (R+1)$.

\subsection{Proof of Lemma~\ref{lem:blockclosed}}
Using the above structure we obtain the following lemma concerning the closed form expression of the iterates. Let us define for any matrix $\bs{A} \in \mathbb{R}^{d\times d}$ and any vector $\bs{b} \in \mathbb{R}^{d}$:
\begin{align*}
&\bs{A}_{(1)} = \{\bs{A}_{ij}: 1\leq i, j \leq R \},\\
&\bs{A}_{(2)} = \{\bs{A}_{ij}: R+1\leq i \leq d, 1\leq j \leq R \},\\
&\bs{b}_{(1)} = \{\bs{b}_{i}: 1\leq i \leq R \},\\
&\bs{b}_{(2)} = \{\bs{b}_{i}: R+1\leq i \leq d \},
\end{align*}
where $R$ is the rank of the data matrix $\bs{D}$ and $d$ is the dimension of the data.

\begin{lemma}\label{lem:blockclosed}
	If $\bs{D}(t)$ is full rank for all $t\geq 0$ and regularizer $\lambda = 0$, then for any $T \geq 0$, the closed form of the iterate $\tilde{\bs{w}}(T)$ admits the following expression:
	\begin{align*}
	&\tilde{\bs{w}}_{(1)}(T) = \bs{A}(T\mathtt{-}1, 0)\tilde{\bsb}_1(0)\\
	&+ \sum_{i=0}^{T-1} \bs{A}(T\mathtt{-}1, i\mathtt{+}1) \eta \tilde{\bs{D}}_{(1)}(i) \bs{\Lambda}^2_{(1)}( \bsb^*_{(1)} + \bs{\Lambda}^{-1}_{(1)} \bs{\zeta}_{(1)}),\\
	&\tilde{\bs{w}}_{(2)}(T) = \bs{B}(T\mathtt{-}1, 0) \tilde{\bsb}_{(1)}(0) + \tilde{\bsb}_{(2)}(0)\\
	&+ \sum_{i=0}^{T-1}\eta \left(\bs{B}(T\mathtt{-}1, i\mathtt{+}1)\tilde{\bs{D}}_{(1)}(i) \mathtt{+} \tilde{\bs{D}}_{(2)}(i) \right) \times \\
	&\times \bs{\Lambda}^2_{(1)}( \bsb^*_{(1)} + \bs{\Lambda}^{-1}_{(1)} \bs{\zeta}_{(1)}),
	\end{align*} 
	where for all $t_2 \geq t_1 \geq 0$,
	\begin{align*}
	&\bs{A}(t_2, t_1) = \prod_{i=t_1}^{t_2}\left(\bs{I} - \eta \tilde{\bs{D}}_{(1)}(i) \bs{\Lambda}_{(1)}^2 \right)\\
	&\bs{B}(t_2, t_1) \mathtt{=} - \eta\tilde{\bs{D}}_{(2)}(t_1)\bs{\Lambda}_{(1)}^2 \\
	&\hspace{1.7cm}\mathtt{-} \eta\sum_{i=t_1\mathtt{+}1}^{t_2} \tilde{\bs{D}}_{(2)}(i)\bs{\Lambda}_{(1)}^2\bs{A}(i\mathtt{-}1, t_1)
	\end{align*}
\end{lemma}
In the above lemma, the vector $\tilde{\bs{w}}_{(1)}(T)$ represents the in-span component of the iterate, where as $\tilde{\bs{w}}_{(2)}(T)$ represents the out-of-span component of the iterate. We make an important observation in the complex expression in Lemma~\ref{lem:blockclosed} that for appropriate choice of $\eta$, we have $\max |\lambda(\bs{A}(t_2, t_1))| < 1$  for all $t_2\geq t_1 \geq 0$. This is true because, even though  $\lambda_{min}(\bs{\Lambda}^2) = 0$,  when only the $R\times R$ submatrix  $\bs{\Lambda}_{(1)}$ is considered, we have $\lambda_{min}(\bs{\Lambda}_{(1)}^2) > 0$. Using this result we prove the convergence of in-span component.  
We will use the following equations regarding  the product two specific block matrices. 
\begin{align}\label{eq:blockprod}
\begin{bmatrix}
\bs{A}_1 & \bs{0}\\
\bs{B}_1 & \bs{I}
\end{bmatrix}
\begin{bmatrix}
\bs{A}_2 & \bs{0}\\
\bs{B}_2 & \bs{I}
\end{bmatrix}
= \begin{bmatrix}
\bs{A}_1 \bs{A}_2  & \bs{0}\\
\bs{B}_1 \bs{A}_2 + \bs{B}_2  & \bs{I}
\end{bmatrix}, 
\hspace{1cm}
\begin{bmatrix}
\bs{A}_1 & \bs{0}\\
\bs{B}_1 & \bs{C}_1
\end{bmatrix}
\begin{bmatrix}
\bs{A}_2 & \bs{0}\\
\bs{0} & \bs{0}
\end{bmatrix}
= \begin{bmatrix}
\bs{A}_1 \bs{A}_2  & \bs{0}\\
\bs{B}_1 \bs{A}_2  & \bs{0}
\end{bmatrix}
\end{align}
Firstly, we obtain the block structure shown in the paper. 
\begin{align*}
\left(\bs{I} - \eta \tilde{\bs{D}}(i) \bs{\Lambda}^2\right) = \bs{I} - \eta 
\begin{bmatrix}
\tilde{\bs{D}}_{(1)}(i) & \tilde{\bs{D}}_{(2)}(i)\\
\tilde{\bs{D}}_{(2)}(i) & \tilde{\bs{D}}_{(3)}(i)\\
\end{bmatrix}
\begin{bmatrix}
\bs{\Lambda}^2_{(1)} & \bs{0}\\
\bs{0} & \bs{0}
\end{bmatrix}
= \begin{bmatrix}
\left(I - \eta \tilde{\bs{D}}_{(1)}(i)\bs{\Lambda}^2[1] \right) & \bs{0}_{R\times (d-R)}\\
- \eta \tilde{\bs{D}}_{(2)}(i) \bs{\Lambda}^2[1] & \bs{I}_{(d-R)\times (d-R)}
\end{bmatrix}
\end{align*}

The block structure is maintained for the product of these matrices, i.e. for all $t_2 \geq t_1 \geq 0$,
\begin{align*}
&\prod_{i=t_1}^{t_2}\left(I - \eta \tilde{\bs{D}}(i) \bs{\Lambda}^2\right)
= \begin{bmatrix}
\bs{A}(t_2, t_1)& \bs{0}_{R\times (d-R)}\\
\bs{B}(t_2, t_1) & \bs{I}_{(d-R)\times (d-R)}
\end{bmatrix},\\
&\bs{A}(t_2, t_1) = \prod_{i=t_1}^{t_2}\left(I - \eta \tilde{\bs{D}}_{(1)}(i) \bs{\Lambda}_{(1)}^2 \right),
\bs{B}(t_2, t_1) \mathtt{=} - \eta\tilde{\bs{D}}_{(2)}(t_1)\bs{\Lambda}_{(1)}^2 \mathtt{-} \eta\sum_{i=t_1\mathtt{+}1}^{t_2} \tilde{\bs{D}}_{(2)}(i)\bs{\Lambda}_{(1)}^2\bs{A}(i\mathtt{-}1, t_1)
\end{align*}
This can be shown easily using induction and using Equation~\eqref{eq:blockprod}.

Substituting these results in the closed form of the iterates in proposition~\ref{lem:spectralclosed} we obtain 
\begin{align*}
&\tilde{\bsb}(T) = \begin{bmatrix}
\bs{A}(T\mathtt{-}1, 0)\tilde{\bsb}_1(0)\\
\bs{B}(T\mathtt{-}1, 0) \tilde{\bsb}_1(0) + \tilde{\bsb}_{(2)}(0)
\end{bmatrix}  + 
\sum_{i=0}^{T-1} \begin{bmatrix}
\bs{A}(T\mathtt{-}1, i\mathtt{+}1) & \bs{0}\\
\bs{B}(T\mathtt{-}1, i\mathtt{+}1) & \bs{I}
\end{bmatrix}
\begin{bmatrix}
\eta \tilde{\bs{D}}_1(i) \bs{\Lambda}^2_1( \bsb^*_1 + \bs{\Lambda}^{-1}_1 \bs{\zeta}_1)\\
\eta \tilde{\bs{D}}_{(2)}(i) \bs{\Lambda}^2_1( \bsb^*_1  + \bs{\Lambda}^{-1}_1 \bs{\zeta}_1)
\end{bmatrix}
\end{align*}
\paragraph{In-span Component:} Therefore, the component of in the span of data is $\tilde{\bsb}(T)$
\begin{align*}
&\tilde{\bs{w}}_{(1)}(T) = \bs{A}(T\mathtt{-}1, 0)\tilde{\bsb}_1(0) +  \sum_{i=0}^{T-1} \bs{A}(T\mathtt{-}1, i\mathtt{+}1) \eta \tilde{\bs{D}}_1(i) \bs{\Lambda}^2_{(1)}( \bsb^*_{(1)} + \bs{\Lambda}^{-1}_{(1)} \bs{\zeta}_{(1)}),
\end{align*}
Similar to the regularized case, we have for any $\eta \in \left(0,  2\left(\lambda_{max}(\bs{D}(t))(\lambda_{max}^2 (\bs{X}))\right)^{-1}\right)$ the in-span component converges. Further, from the fixed point argument we know that $\bs{v}_r^T \hat{\bsb} =\tilde{\bsb}^*_r + \lambda_r^{-1} \tilde{\bs{\zeta}}_r$. 
\paragraph{Out-of-span Component:} The component outside the span of the data is 
\begin{align*}
\tilde{\bs{w}}_{(2)}(T) &= \bs{B}(T\mathtt{-}1, 0) \tilde{\bsb}_{(1)}(0) + \tilde{\bsb}_{(2)}(0) \\
&+ \sum_{i=0}^{T-1}\eta \left(\bs{B}(T\mathtt{-}1, i\mathtt{+}1)\tilde{\bs{D}}_{(1)}(i) \mathtt{+} \tilde{\bs{D}}_{(2)}(i) \right)\bs{\Lambda}^2_{(1)}
( \bsb^*_{(1)} + \bs{\Lambda}^{-1}_{(1)} \bs{\zeta}_{(1)}).
\end{align*}

\subsection{Proof of Theorem~\ref{thm:unregularized}}
The convergence of the in-span component follows similar to the regularized case. In particular, we observe
$$\lambda_{max}(\bs{I} - \eta \tilde{\bs{D}}_{(1)}(t) \bs{\Lambda}_{(1)}^2) \leq  1 - \eta \lambda_{min}(\tilde{\bs{D}}_{(1)}(t)) \lambda_{min}(\bs{\Lambda}_{(1)}^2) < 1.$$ 
The last inequality is true as 1) $\lambda_{min}(\tilde{\bs{D}}_{(1)}(t)) > 0$ due to the positive definiteness of the matrix $\tilde{\bs{D}}(t)$, and 2) $\lambda_{min}(\bs{\Lambda}_{(1)}^2) > 0$ as it considers only the in-span component (i.e. the top-left $R\times R$ sub-matrix of $\bs{\Lambda}$). On the other hand, we have 
$\lambda_{min}(\bs{I} - \eta \tilde{\bs{D}}_{(1)}(t) \bs{\Lambda}_{(1)}^2) \geq  1 - \eta \lambda_{max}(\tilde{\bs{D}}_{(1)}(t)) \lambda_{max}(\bs{\Lambda}_{(1)}^2)$. Therefore, we obtain $\lambda_{min}(\bs{I} - \eta \tilde{\bs{D}}_{(1)}(t) \bs{\Lambda}_{(1)}^2) > -1$ for any  $0< \eta < 2/ (\lambda_{max}(\tilde{\bs{D}}_{(1)}(t)) \lambda_{max}(\bs{\Lambda}_{(1)}^2))$. 

As  $\tilde{\bs{D}}_{(1)}(t)$ is a principal sub-matrix of $\tilde{\bs{D}}(t)$ for each $t\geq 0$, we have from Cauchy Interlacing Theorem $\lambda_{max}(\tilde{\bs{D}}_{(1)}(t)) \leq \lambda_{max}(\tilde{\bs{D}}(t)) = \lambda_{max}(\bs{D}(t))$. The last equality is due to Proposition~\ref{lem:tildeD}. The characterization of the fixed point follows the same argument as Proposition~\ref{lem:spectralRegularized}.

To prove the second part, we further simplify the out-of-span component using exchange of summation (for finite $T$). Here, we use the convention $\bs{A}(t_1, t_2) = \bs{I}$ for any $t_1 < t_2$.
\begin{align*}
&\tilde{\bs{w}}_{(2)}(T) - \left(\bs{B}(T\mathtt{-}1, 0) \tilde{\bsb}_{(1)}(0) + \tilde{\bsb}_{(2)}(0) \right)\\
&= \sum_{i=0}^{T-1}\eta\left(\tilde{\bs{D}}_{(2)}(i) - \eta\tilde{\bs{D}}_{(2)}(i+1)\bs{\Lambda}_{(1)}^2\tilde{\bs{D}}_{(1)}(i)  \mathtt{-} \right.\\
&\quad \quad\left.\eta\sum_{j=i\mathtt{+}2}^{T-1} \tilde{\bs{D}}_{(2)}(i)\bs{\Lambda}_{(1)}^2\bs{A}(j\mathtt{-}1, i+1)\tilde{\bs{D}}_{(1)}(i)  \right) \bs{\Lambda}^2_{(1)}
( \bsb^*_{(1)} + \bs{\Lambda}^{-1}_{(1)} \bs{\zeta}_{(1)}).\\
&=\left(\sum_{i=0}^{T-1}\eta\tilde{\bs{D}}_{(2)}(i) - \eta^2 \sum_{i=0}^{T-1}\sum_{j=i+1}^{T-1}  \tilde{\bs{D}}_{(2)}(j)\bs{\Lambda}_{(1)}^2\bs{A}(j\mathtt{-}1, i+1)\tilde{\bs{D}}_{(1)}(i)\right) \bs{\Lambda}^2_{(1)}
( \bsb^*_{(1)} + \bs{\Lambda}^{-1}_{(1)} \bs{\zeta}_{(1)}). \\
& = \left(\sum_{i=0}^{T-1} \eta \tilde{\bs{D}}_{(2)}(i) \left( \bs{I}  - \eta \bs{\Lambda}_{(1)}^2 \sum_{j=0}^{i-1} \bs{A}(i\mathtt{-}1, j+1)\tilde{\bs{D}}_{(1)}(j) \right)\right) \bs{\Lambda}^2_{(1)}
( \bsb^*_{(1)} + \bs{\Lambda}^{-1}_{(1)} \bs{\zeta}_{(1)}),\\
&  = \left(\sum_{i=0}^{T-1} \eta \tilde{\bs{D}}_{(2)}(i) \left( \bs{\Lambda}^2_{(1)}
( \bsb^*_{(1)} + \bs{\Lambda}^{-1}_{(1)} \bs{\zeta}_{(1)})  - \bs{\Lambda}_{(1)}^2 \tilde{\bs{w}}_{(1)}(i) + \bs{\Lambda}_{(1)}^2 \bs{A}(i\mathtt{-}1, 0)\tilde{\bsb}_1(0) \right)\right),\\
&  = \sum_{i=0}^{T-1} \eta \tilde{\bs{D}}_{(2)}(i) \bs{\Lambda}^2_{(1)}
\left( (\bsb^*_{(1)} + \bs{\Lambda}^{-1}_{(1)} \bs{\zeta}_{(1)})  - \tilde{\bs{w}}_{(1)}(i)\right) 
+ \sum_{i=0}^{T-1} \eta \tilde{\bs{D}}_{(2)}(i)\bs{\Lambda}_{(1)}^2 
\bs{A}(i\mathtt{-}1, 0)\tilde{\bsb}_1(0).
\end{align*}

Therefore, we have 
\begin{align*}
&\|\tilde{\bs{w}}_{(2)}(T) - \tilde{\bsb}_{(2)}(0)\|_2 \\
&\leq \sum_{i=0}^{T-1} \eta \tilde{\bs{D}}_{(2)}(i) \bs{\Lambda}^2_{(1)}
\left( (\bsb^*_{(1)} + \bs{\Lambda}^{-1}_{(1)} \bs{\zeta}_{(1)})  - \tilde{\bs{w}}_{(1)}(i)\right) + \left(\bs{B}(T\mathtt{-}1, 0) + \sum_{i=0}^{T-1} \eta \tilde{\bs{D}}_{(2)}(i)\bs{\Lambda}_{(1)}^2 
\bs{A}(i\mathtt{-}1, 0)\right)\tilde{\bsb}_1(0) \\
&= \sum_{i=0}^{T-1} \eta \tilde{\bs{D}}_{(2)}(i) \bs{\Lambda}^2_{(1)}
\left( (\bsb^*_{(1)} + \bs{\Lambda}^{-1}_{(1)} \bs{\zeta}_{(1)})  - \tilde{\bs{w}}_{(1)}(i)\right) 
- \eta\tilde{\bs{D}}_{(2)}(0)\bs{\Lambda}_{(1)}^2\tilde{\bsb}_1(0)
\end{align*}

We have $|\lambda|_{\max} = \sup_t |\lambda_{\max}(I- \eta\tilde{\bs{D}_1(t)}\bs{\Lambda}_{(1)}^2)| < 1$ due to appropriate choice of $\eta$.  Also, by assumption of the theorem  we have for some $\alpha \geq 0$, $\beta \geq 0$, $\alpha + \beta > 1$, for some universal constants $ 0 < c_{conv}, c_{\lambda} <\infty$,  and for all $t\geq 0$:

(i) the out-of-span pre-conditioner matrix decaying as $O(1/t^\alpha)$ for some $\alpha \geq 0$, i.e. $|\lambda_{\max}(\tilde{\bs{D}}_{(2)}(i))| = \tfrac{c_{\lambda}}{(t+1)^\alpha}$, 
and 

(ii) the convergence rate of the in-span component is $O(1/t^\beta)$ with iteration $t$ for some $\beta > 0$, i.e. $\|(\bsb^*_{(1)} + \bs{\Lambda}^{-1}_{(1)} \bs{\zeta}_{(1)})  - \tilde{\bs{w}}_{(1)}(i)\|_2 \leq \tfrac{c_{conv}}{(t+1)^\beta}$.

For the first term we have,
\begin{align*}
&\|\sum_{i=0}^{T-1} \eta \tilde{\bs{D}}_{(2)}(i) \bs{\Lambda}^2_{(1)}
\left( (\bsb^*_{(1)} + \bs{\Lambda}^{-1}_{(1)} \bs{\zeta}_{(1)})  - \tilde{\bs{w}}_{(1)}(i)\right)\|_2\\
&\leq c_{\lambda} \sum_{i=0}^{T-1}\lambda_{\max}(\tilde{\bs{D}}_{(2)}(i))\|(\bsb^*_{(1)} + \bs{\Lambda}^{-1}_{(1)} \bs{\zeta}_{(1)})  - \tilde{\bs{w}}_{(1)}(i) \|_2\\
&\leq c_{\lambda} c_{conv} \sum_{i=1}^{T-1} \tfrac{1}{(i+1)^{(\alpha+\beta)}}
\leq \tfrac{ c_{\lambda} c_{conv}}{\alpha+\beta-1} \left(1 - \tfrac{1}{(i+1)^{(\alpha+\beta-1)}}\right).
\end{align*}
Therefore, the first term saturates to a value at most $\tfrac{c_{\lambda} c_{conv}}{\alpha+\beta-1}$. 

For the second term we have,
\begin{align*}
&\|\eta\tilde{\bs{D}}_{(2)}(0)\bs{\Lambda}_{(1)}^2\tilde{\bsb}_1(0)\|_2 
\leq  \eta \lambda_{\max}(\tilde{\bs{D}}_{(2)}(0)) \lambda^2_{\max}(\bs{\Lambda}_{(1)})\|\tilde{\bsb}_1(0)\|_2 
\end{align*}



\subsection{Proof of Proposition~\ref{lem:goodAdaptive}}

\begin{proposition}\label{lem:goodAdaptive}
	The following pre-conditioner matrices have $\tilde{\bs{D}}_{(2)}(t) = \bs{0}$.
	\begin{enumerate}
		\item  $\bs{D}(t)= \bs{I}$, i.e. gradient descent,
		\item $\bs{D}(t) = (\bs{X}^T \bs{X} + \epsilon \bs{I})^{-1}$ for all $t \geq 0$.
	\end{enumerate}
\end{proposition}

We have $(\bs{X}^T \bs{X} + \epsilon \bs{I})^{-1} = \sum_{r = 1}^{R} (\lambda_r^2 + \epsilon)^{-1}\bs{v}_r \bs{v}^T_r +  \sum_{r = R+1}^{d} \epsilon^{-1}\bs{v}_r \bs{v}^T_r$. Further, $\bs{I} = \sum_{r = 1}^{d} bs{v}_r \bs{v}^T_r$. So the proposition is true.

\subsection{Proof of Lemma \ref{lemma:adagaradvariantproof}}

We will prove this using induction. 
Let $Q =  diag(|X^T y|)$
We will show that 
\begin{align*}
w_k = \lambda_k Q^{-1} sign(X^T y)
\end{align*}
for some $\lambda_k$. $w_0 = 0$ is satsified for $\lambda_0=0$ and so the base case is trivially true. 
\begin{align*}
g_k &= X^T (X w_k - y)\\
&=  \lambda_k X^T X Q^{-1} sign(X^T y) - X^T y\\
&= (\lambda_k c - 1)X^T y
\end{align*}
where the last inequality follows from $w_k = \lambda_k Q^{-1} sign(X^T y)$.3
\begin{align*}
H_k = diag(\sum_{s=1}^n g_s \cdot g_s) = \nu_k diag(|X^T y|^2) = \nu_k Q^2
\end{align*}
\begin{align*}
w_{k+1} &= w_{k} - \alpha_k H_k^{-1} X^T (X w_k - y) \\
&= w_k - \alpha_k H_k^{-1} X^T X w_k  + \alpha_k H_k^{-1} X^T y \\
&= \lambda_k Q^{-1} sign(X^T y) - \lambda_k \alpha_k H_k^{-1} X^T X  Q^{-1}  X^T y + \alpha_k H_k^{-1} X^T y \\
&= \lambda_k Q^{-1} sign(X^T y)  - \lambda_k \alpha_k c H_k^{-1} X^T y + \alpha_k H_k^{-1} X^T y\\
&= \left( \lambda_k - \frac{\lambda_k \alpha_k c}{\nu_k} + \frac{\alpha_k}{\nu_k}\right) Q^{-1} sign(X^T y) \\
&= \lambda_{k+1} Q^{-1}  sign(X^T y)
\end{align*}


\section{Additional Experiments}

\subsection{Generalization with respect to test accuracy for over-parameterized linear regression}{\label{sec:counter}} 

In the first set of experiments, we showed how adaptive methods converging to a different solution might lead to solutions farther from $\bsb^*_b$, i.e. with higher L2-norm . Thus, the pre-conditioner matrices satisfying $\bs{D}_{ij}(t) = 0 , \bs{D}(t)\succ 0$ have different generalization than their gradient based counterparts. In this section, we empirically demonstrate that pre-conditioner matrices of the form: $\bs{D}_{ij}(t) = 0 , \bs{D}(t)\succ 0$ can guarantee better generalization than gradient based methods depending on the problem in hand. As a direct consequence of this, we show that solutions with a minimum norm should not be used as a yardstick to guarantee good generalization. 

We alter the previous counterexample in \citep{wilson2017marginal} by slightly changing the problem setting: at first, we reduce the margin between the two classes; the case where we increase the margin is provided in the Appendix.
We \emph{empirically} show that gradient-descent methods fail to generalize as well as adaptive methods --with a slightly different $D_k$ than AdaGrad. 
In particular, for the responses, we consider two classes $y_i \in \{\pm \ell\}$ for some $\ell \in (0, 1)$; \emph{i.e.}, we consider a smaller margin between the two classes.
$\ell$ can take different values, and still we get the same performance, as we show in the experiments below.
\begin{small}
	\begin{align} 
	\left(x_i\right)_j = 
	\begin{cases}
	y_i \ell, & \!\!j = 1, \\
	1, & \!\!j = 2, 3,\\
	1, &  \!\!j=4+ 5(i-1), \\
	0, & \!\!\text{otherwise}.
	\end{cases}  ~~~~~ \text{if} ~ y_i = 1, 
	\qquad \left(x_i\right)_j = 
	\begin{cases}
	y_i\ell, & \!\!j = 1, \\
	1, & \!\!j = 2, 3, \\
	1, & \!\!j=  4+ 5(i-1),\\
	& \hspace{0.3cm}  \cdots, 8+ 5(i-1), \\
	0, & \text{otherwise}.
	\end{cases} ~~~ \text{if} ~ y_i = -1. \label{eq:features}
	\end{align}
\end{small}

Given this generative model, we construct $n$ samples $\{y_i, x_i\}_{i=1}^n$, and set $d = 6n$, for different $n$ values.
We compare two simple algorithms: $\pmb{i)}$ the plain gradient descent,
for $\eta = \sfrac{1}{\lambda_1(\bs{X}^\top \bs{X})}$; 
$\pmb{ii)}$ the recursion $\bs{w}(t+1) = \bs{w}(t) - \eta \bs{D}(t) \bs{X}^\top \left(\bs{X} \bs{w}(t) - \bs{y}\right)$, where $\eta$ is set as above, and $\bs{D}(t)$ follows the rule:
\begin{small}
	\begin{align}\label{eq:AdaGradvariant}
	\bs{D}(t) &= \texttt{diag}\left( 1 / \left(\sum_{j = t - J}^t \nabla f(\bs{w}(j)) \odot\nabla f(\bs{w}(j)) + \varepsilon\right)^2\right) \notag\\
	&\succ 0, \quad \quad \text{for some} ~\varepsilon > 0,	~~\text{and}~~ J < t \in \mathbb{N}_{+} 
	\end{align}
\end{small}
Observe that $\bs{D}(t)$ uses the dot product of gradients, \emph{squared}. 
A variant of this preconditioner is found in \cite{mukkamala2017variants}; however our purpose is not to recommend a particular preconditioner but to show that \emph{there are $D_k$ that lead to better performance than the minimum norm solution. }
We denote as $w^{\text{ada}}$,  $w^{\text{adam}}$ and $w^{\text{GD}}$ the estimates of the adam, adagrad variant and simple gradient descent, respectively.

The experiment obeys the following steps: $\pmb{i)}$ we train both gradient and adaptive gradient methods on the same training set, 
$\pmb{ii)}$ we test models on new data $\{y_i^{\text{test}}, x_i^{\text{test}}\}_{i = 1}^Q$.

We define performance in terms of the classification error: for a new sample $\{y_i^{\text{test}}, x_i^{\text{test}}\}$ and given $w^{\text{ada}}$, $w^{\text{adam}}$ and $w^{\text{GD}}$, the only features that are non-zeros in both $x_i^{\text{test}}$ and $w$'s are the first 3 entries \cite[pp. 5]{wilson2017marginal}. 
This is due to the fact that, for gradient descent and given the structure in $\bs{X}$, only these 3 features 
affects the performance of gradient descent.
Thus, the decision rules for both algorithms are:
\begin{small}
	\begin{align*}
	\widehat{y}_i^{~\text{ada}} &= 
	\texttt{quant}_{\ell}\left( w^{\text{ada}}_1 \cdot y_i^{\text{test}} + w^{\text{ada}}_2 + w^{\text{ada}}_3 \right), \\ ~\widehat{y}_i^{~\text{GD}} &= \texttt{quant}_{\ell}\left( w^{\text{GD}}_1 \cdot y_i^{\text{test}} + w^{\text{GD}}_2 + w^{\text{GD}}_3 \right),\\
	~\widehat{y}_i^{~\text{adam}} &= \texttt{quant}_{\ell}\left( w^{\text{adam}}_1 \cdot y_i^{\text{test}} + w^{\text{adam}}_2 + w^{\text{adam}}_3 \right),
	\end{align*}
\end{small}
where $\texttt{quant}_{\ell}(\alpha)$ finds the nearest point w.r.t. $\{\pm \ell\}$.
With this example, our aim is to show that adaptive methods lead to models that have better generalization than gradient descent.

Table 2 summarizes the empirical findings. 
In order to cover a wider range of settings, we consider $n = [10, ~50, ~100]$ and set $d = 6n$, as dictated by \cite{wilson2017marginal}.
We generate $\bs{X}$ as above, where instances in the positive class, $y_i \in +\ell$, are generated with probability $p = 7/8$; the cases where $p = 5/8$ and $p = 3/8$ are provided in the appendix, and also convey the same message as in Table \ref{table:1a_main}.  Further details on the experiments are provided in the Appendix.

The proposed AdaGrad variant described in equation \ref{eq:AdaGradvariant} falls under the broad class of adaptive algorithms with $D_k$. 
However, for the counter example in \cite[pp. 5]{wilson2017marginal}, the AdaGrad variant neither satisfies the convergence guarantees of Lemma 3.1 there, nor does it converge to the minimum norm solution evidenced by its norm in Table \ref{table:1a_main}.
To buttress our claim that the AdaGrad variant in \eqref{eq:AdaGradvariant} converges to a solution different than that of minimum norm (which is the case for plain gradient descent), we provide the following proposition for a specific class of problems\footnote{Not the problem proposed in the counter-example 1 on pg 5.}; the proof is provided in Appendix.

\begin{proposition} \label{lemma:adagaradvariantproof}
	Suppose $\bs{X}^\top \bs{y}$ has no zero components. Define $\bs{D} = \texttt{diag}(|\bs{X}^\top\bs{y}|^{3})$  and assume there exists a scalar $c$ such that $\bs{X} \bs{D}^{-1} \texttt{sign}(\bs{X}^\top \bs{y}) = c \bs{y}$. 	Then, when initialized at 0, the AdaGrad variant in \eqref{eq:AdaGradvariant} converges to the unique solution $\bs{w} \propto \bs{D}^{-1} \texttt{sign}(\bs{X}^\top \bs{y})$.
\end{proposition}
\vspace{0.1cm}

\subsection{More details and experiments for the counter-example}

The simulation is completed as follows: 
For each setting $(n, p, J)$, we generate 100 different instances for $(X, y)$, and for each instance we compute the solutions from gradient descent, AdaGrad variant and Adam (RMSprop is included in the Appendix) and the minimum norm solution $w_{\text{mn}}$.  
In the appendix, we have the above table with the Adagrad variant that normalizes the final solution $\widehat{w}$ (Table \ref{table:1a_main_extra}) before calculating the distance w.r.t. the minimum norm solution: we observed that this step did not improve or worsen the performance, compared to the unnormalized solution.
\emph{This further indicates that there is an infinite collection of solutions --with different magnitudes-- that lead to better performance than plain gradient descent; thus our findings are not a pathological example where adaptive methods work better.}

We record $\|\widehat{w}- w_{\text{mn}}\|_2$, where $\widehat{w}$ represents the corresponding solutions obtained by the algorithms in the comparison list. 
For each $(X, y)$ instance, we further generate $\{y_i^{\text{test}}, x_i^{\text{test}}\}_{i = 1}^{100}$, and we evaluate the performance of both models on predicting $y_i^{\text{test}}$, $\forall i$. 

Table \ref{table:1a_main} shows that gradient descent converges to the minimum norm solution, in contrast to the adaptive methods. 
This justifies the fact that the adaptive gradient methods (including the proposed adagrad variant) converge to a different solution than the minimum norm solution.
Nevertheless, the accuracy on \emph{unseen} data is higher in the adaptive methods (both our proposed AdaGrad variant and in most instances, Adam), than the plain gradient descent, when $\ell$ is small: the adaptive method successfully identifies the correct class, while gradient descent only predicts one class (the positive class; this is justified by the fact that the accuracy obtained is approximately close to $p$, as $n$ increases). 
We first provide the same table in Table \ref{table:1a_main}  but with unnormalized values for distances with respect to Adagrad variant.

\begin{table}[!htbp]
	\centering
	\caption{Prediction accuracy and distances from the minimum norm solution for plain gradient descent and adaptive gradient descent methods. We set $p = 7/8$ and $J=10$, as in the main text. The adaptive method uses $D_k$ according to \eqref{eq:AdaGradvariant}. The distances shown are median values out of 100 different realizations for each setting; the accuracies are obtained by testing $10^4$ predictions on unseen data.} \label{table:1a_main_extra11} 
	\vspace{0.3cm}
	\begin{footnotesize}
		\begin{tabular}{c c c c c c c c c c c}
			\toprule
			\phantom{1} & & \phantom{3} & & \phantom{5} & & Gradient Descent & &  AdaGrad variant & & Adam \\
			\midrule
			\multirow{8}{*}{$n = 10$} & & \multirow{2}{*}{$\ell = 1/32$} & & Acc. (\%) & &63 & &\textbf{100}  & &91   \\ 
			\phantom{1}		& & 		\phantom{3}		& & $\|\widehat{w} - w_{\text{mn}}\|_2$ & &$1.015\cdot 10^{-16}$ & &$0.9911$  & &$0.1007$   \\
			\cmidrule{7-11}
			& & \multirow{2}{*}{$\ell = 1/16$} & & Acc. (\%)  & &53 & &\textbf{100}  & &87 \\ 
			& & 						& & $\|\widehat{w} - w_{\text{mn}}\|_2$ & &$1.7401\cdot 10^{-16}$ & &$0.9263$   & &$0.0864$  \\
			\cmidrule{7-11}
			& & \multirow{2}{*}{$\ell = 1/8$} & & Acc. (\%)  & &58 & &\textbf{99}  & &84 \\ 
			& & 						& & $\|\widehat{w} - w_{\text{mn}}\|_2$ & &$4.08\cdot 10^{-16}$ & &$0.8179$  & &$0.0764$ \\
			\midrule
			\multirow{8}{*}{$n = 50$} & & \multirow{2}{*}{$\ell = 1/32$} & & Acc. (\%) & &77 & &\textbf{100} & &88 \\
			\phantom{1}		& & 		\phantom{3}		& & $\|\widehat{w} - w_{\text{mn}}\|_2$ & &$4.729\cdot 10^{-15}$ & &$0.8893$   & &$0.0271$\\
			\cmidrule{7-11}
			& & \multirow{2}{*}{$\ell = 1/16$} & & Acc. (\%)& &80 & &\textbf{100}  & &89\\ 
			& & 						& & $\|\widehat{w} - w_{\text{mn}}\|_2$& &$6.9197\cdot 10^{-15}$ & &$0.7929$   & &$0.06281$\\
			\cmidrule{7-11}
			& & \multirow{2}{*}{$\ell = 1/8$} & & Acc. (\%) & &91 & &\textbf{100}   & &89 \\
			& & 						& & $\|\widehat{w} - w_{\text{mn}}\|_2$ & &$9.7170\cdot 10^{-15}$ & &$0.6639$   & &$0.1767$ \\
			\midrule
			\multirow{8}{*}{$n = 100$} & & \multirow{2}{*}{$\ell = 1/32$} & & Acc. (\%)& &85 & &\textbf{100}  & &95 \\
			\phantom{1}		& & 		\phantom{3}		& & $\|\widehat{w} - w_{\text{mn}}\|_2$ & &$4.975\cdot 10^{-9}$ & &$0.8463$   &&$0.0344$ \\
			\cmidrule{7-11}
			& & \multirow{2}{*}{$\ell = 1/16$} & & Acc. (\%) & &83 & &\textbf{100}  & &76 \\ 
			& & 						& & $\|\widehat{w} - w_{\text{mn}}\|_2$ & &$2.5420\cdot 10^{-9}$ & &$0.7217$   & &$0.1020$\\
			\cmidrule{7-11}
			& & \multirow{2}{*}{$\ell = 1/8$} & & Acc. (\%) & &\textbf{100} & &\textbf{100}  & &90\\ 
			& & 						& & $\|\widehat{w} - w_{\text{mn}}\|_2$& &$1.5572\cdot 10^{-11}$ & &$0.6289$  & &0.3306\\
			\bottomrule
		\end{tabular}
	\end{footnotesize}
\end{table}

Here, we provide further results on the counterexample in Subsubsection \ref{sec:counter}.
Tables \ref{table:counterex11} and \ref{table:1a_main_extra11} contains results for $J = 10$: the purpose of these tables is to show that even if we change the memory use footprint of the AdaGrad variant---by storing fewer or more gradients to compute $D_k$ in \eqref{eq:AdaGradvariant}---the results are the same: the AdaGrad variant consistently converges to a solution different than the minimum norm solution, while being more accurate than the latter for small values of $\ell$ (\emph{i.e.}, smaller margin between the two classes).


Plain gradient descent methods provably need to rely on the first elements to decide; using the same rule for adaptive methods\footnote{We note that using only the three elements in adaptive methods is not backed up by theory since it assumes that the training and test datasets have no overlap. We include this in comparison for completeness.}. 
The remaining subsection considers the case where we decide based on the $y = \texttt{sign}(x^\top w)$ rule, where $w$ is the complete learned model. 
As we show empirically, more often than not adaptive methods outperform plain gradient methods.

Observing the performance of various optimization techniques for different values of $n$, $p$ and $\ell$, 
we observed that the best performances are obtained when the dataset is highly imbalanced irrespective of the optimization algorithm chosen. 
When the data is (almost) balanced, it is difficult to comment on how the performance of these algorithms is affected by variations in the levels $\ell$ and probability $p$.

\graphicspath{{../Code/Fullproductplots/}}

\subsection{Deep Learning}

In this section, we will extend the experiments to over-parameterized and under-parameterized neural networks without regularization.  We begin with a detailed description of the datasets and the architectures we use along with comprehensive set of experiments with hyperparameter tuning. 

\begin{table}[!h]
	\caption{Summary of the datasets and the architectures used for experiments. CNN stands for convolutional neural network, FF stands for feed forward network. More details are given in the main text.}
	\vspace{0.3cm}
	\centering
	\begin{tabular}{c c c c c c c}	
		\toprule
		Name & & Network type & & Dataset \\
		\cmidrule{1-1} \cmidrule{3-3} \cmidrule{5-5} 
		M1-UP & & Shallow CNN + FFN & & MNIST\\
		M1-OP & & Shallow CNN + FFN & & MNIST\\
		C1-UP & & Shallow CNN + FFN & & CIFAR-10 \\
		C1-OP & & ResNet18  & & CIFAR-10 \\
		C2-OP & & PreActResNet18 & & CIFAR-100 \\
		C3-OP & & MobileNet & & CIFAR-100 \\
		C4-OP & & MobileNetV2 & & CIFAR-100 \\
		C5-OP & & GoogleNet& & CIFAR-100 \\
		\bottomrule
	\end{tabular} \label{table:datasets}
\end{table}



\begin{figure*}[!htb]
	\centering
	\includegraphics[width=0.24\textwidth]{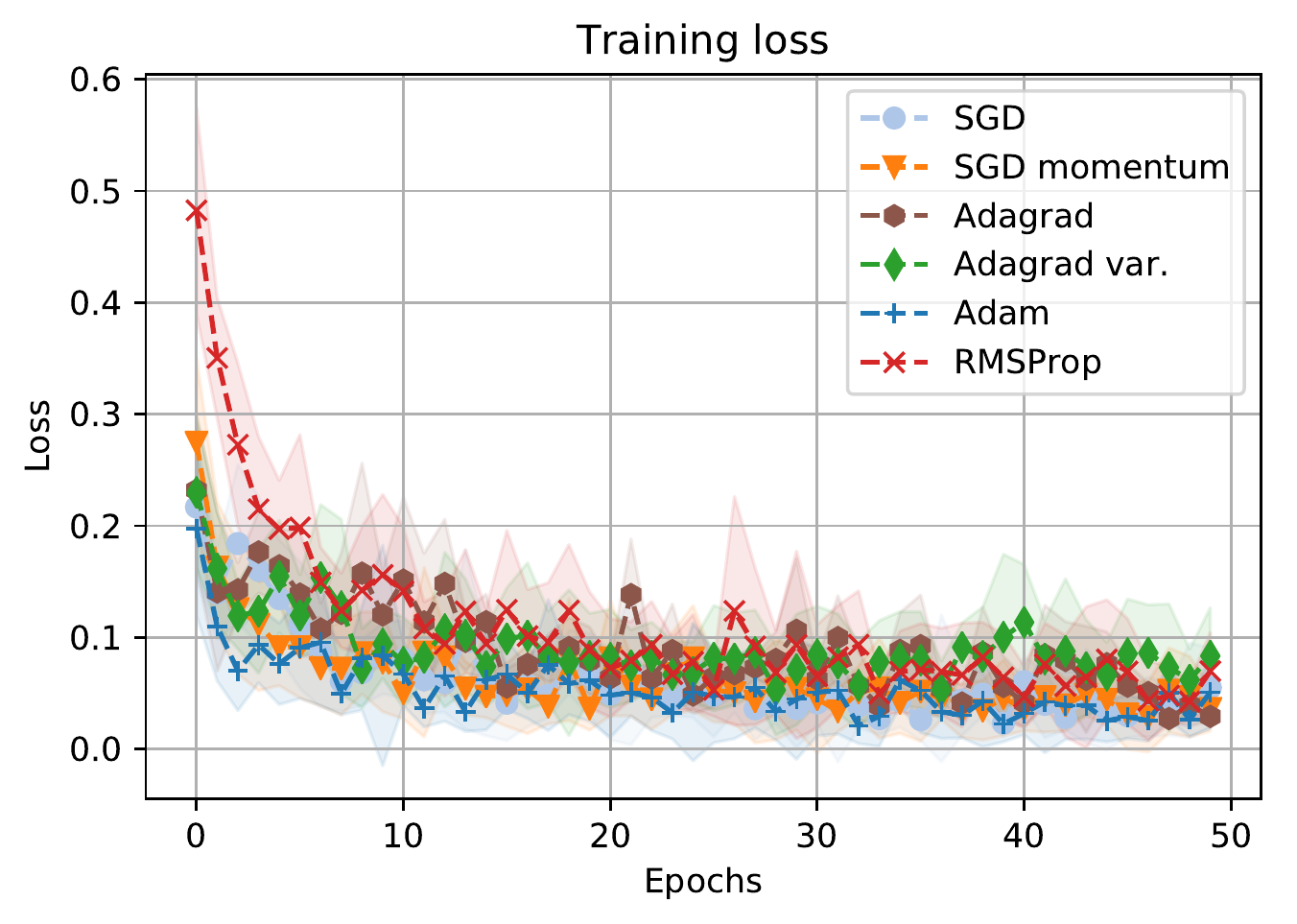}
	\includegraphics[width=0.24\textwidth]{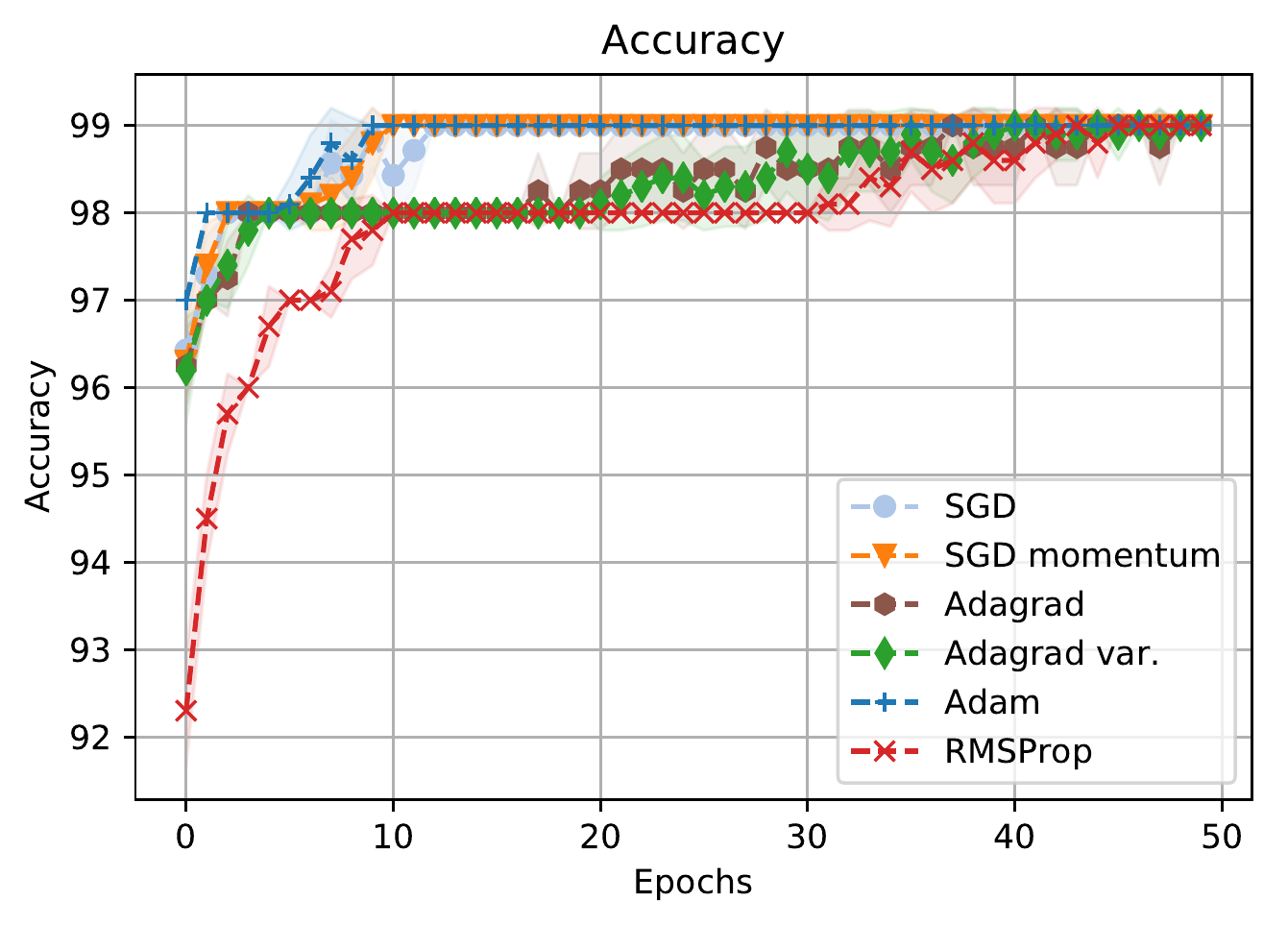}	
	\includegraphics[width=0.24\textwidth]{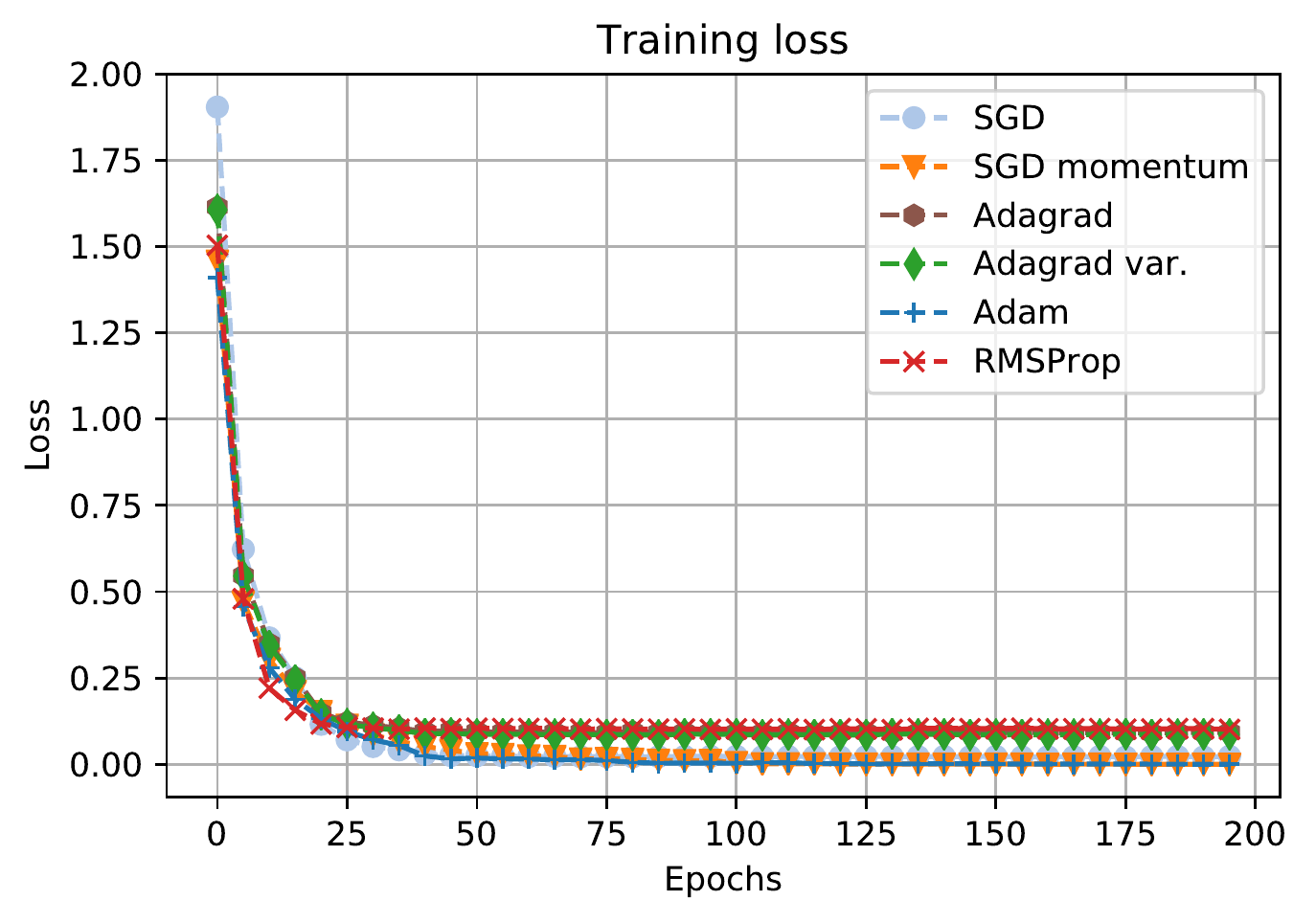}  
	\includegraphics[width=0.24\textwidth]{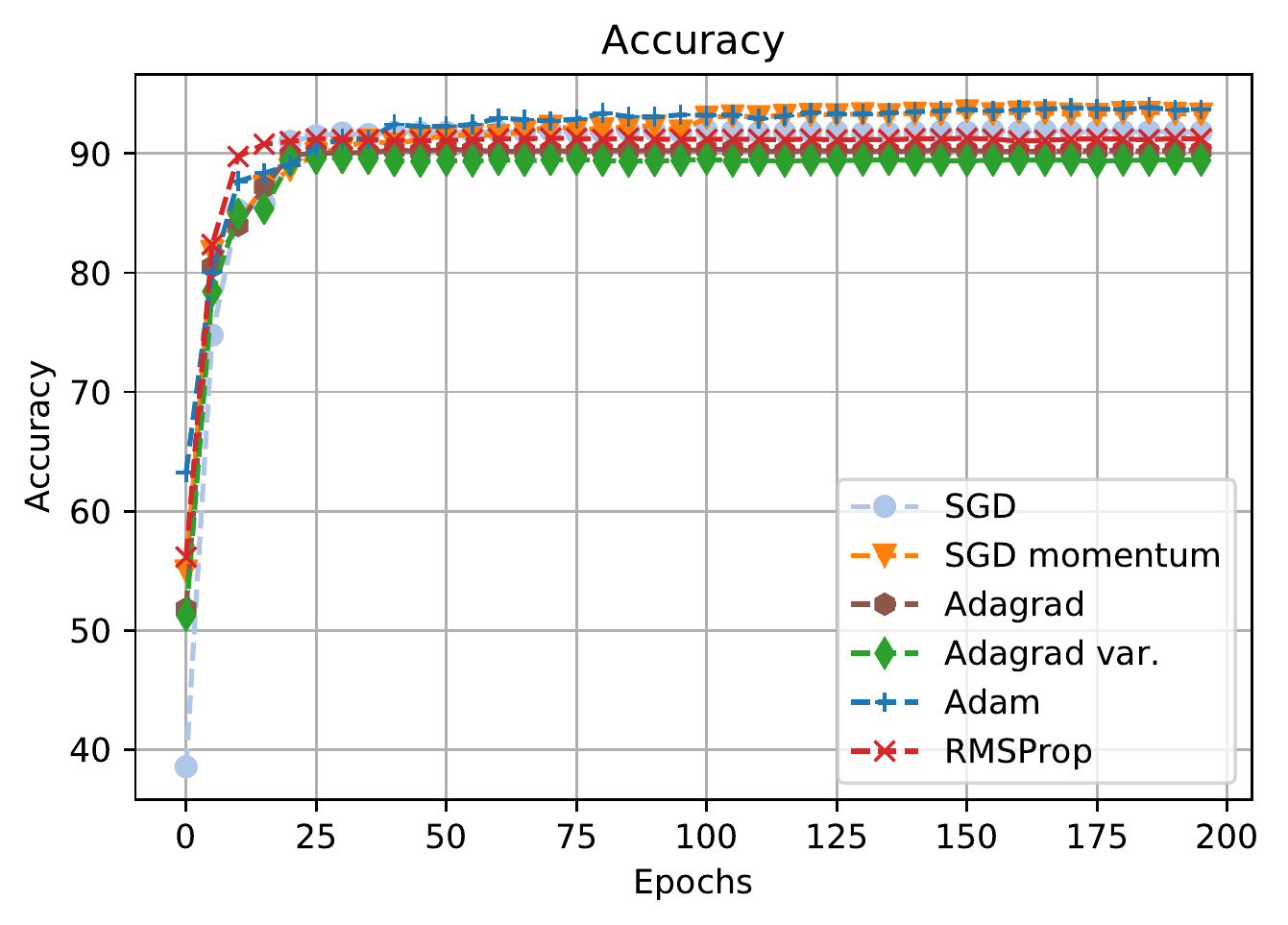} 
	\caption{Accuracy results on unseen data, for different NN architectures and datasets. Over-parameterized problems. \emph{Left two panels:} Accuracy and training loss for MNIST; \emph{Right two panels:} Accuracy and training loss for CIFAR10.} \label{fig:00}
\end{figure*}

\paragraph{MNIST dataset and the M1 architecture.}
Each experiment for M1 is simulated over 50 epochs and 10 runs for both under- and over-parameterized settings. 
Both the MNIST architectures consisted of two convolutional layers (the second one with dropouts  \cite{srivastava2014dropout}) followed by two fully connected layers. 
The primary difference between the M1-OP ($\sim73$K parameters) and M1-UP ($\sim21$K parameters) architectures was the number of channels in the convolutional networks and $\#$ of nodes in the last fully connected hidden layer. 

Figure \ref{fig:00}, left two columns, reports the results over 10 Monte-Carlo realizations.
Top row corresponds to the M1-UP case; bottom row to the M1-OP case.
We plot both training errors and the accuracy results on unseen data. 
For the M1-UP case, despite the grid search, observe that AdaGrad (and its variant) do not perform as well as the rest of the algorithms. 
Nevertheless, adaptive methods (such as Adam and RMSProp) perform similarly to simple SGD variants, supporting our conjecture that each algorithm requires a different configuration, but still can converge to a good local point; also that adaptive methods require the same (if not more) tuning.
For the M1-OP case, SGD momentum performs less favorably compared to plain SGD, and we conjecture that this is due to non-optimal tuning. 
In this case, all adaptive methods perform similarly to SGD.

\paragraph{CIFAR10 dataset and the C1 architecture.}
For C1, C1-UP is trained over $350$ epochs, while C1-OP was trained over $200$ epochs.  
The under-parameterized setting is on-purpose tweaked to ensure that we have fewer parameters than examples ($\sim43$K parameters), and slightly deviates from \citep{mcdonnell2015enhanced}; our generalization guarantees ($\sim76\%$) are in conjunction with the attained test accuracy levels. 
Similarly, for the C1-OP case, we implement a Resnet \citep{he2016deep} + dropout architecture ($\sim0.25$ million parameters) 
Adam and RMSProp achieves the best performance than their non-adaptive counterparts for both the under-parameterized and over-parameterized settings.

Figure \ref{fig:00}, right panel, follows the same pattern with the MNIST data; it reports the results over 10 Monte-Carlo realizations.
Again, we observe that AdaGrad methods do not perform as well as the rest of the algorithms. 
Nevertheless, adaptive methods (such as Adam and RMSProp) perform similarly to simple SGD variants. Further experiments on CIFAR-100 for different architecture are provided in the Appendix.

\paragraph{CIFAR100 and other deep architectures (C\{2-5\}-OP).}

In this experiment, we focus only on the over-parameterized case: DNNs are usually designed over-parameterized in practice, with ever growing number of layers, and, eventually, a larger number of parameters \citep{telgarsky2016benefits}.
We again completed 10 runs for each of the set up we considered.
C2-OP corresponds to PreActResNet18 from \citep{he2016identityb}, C3-OP corresponds to MobileNet from \citep{howard2017mobilenets}, C4-OP is MobileNetV2 from \citep{sandler2018inverted}, and C5-OP is GoogleNet from \citep{szegedy2015going}.
The results are depicted in Figure \ref{fig:01}.
After a similar hyper-parameter tuning phase, we selected the best choices among the parameters tested.
The results show no clear winner once again, which overall support our claims: \emph{the superiority depends on the problem/data at hand; also, all algorithms require fine tuning to achieve their best performance.}
We note that a more comprehensive reasoning requires multiple runs for each network, as other hyper-parameters (such as initialization) might play significant role in closing the gap between different algorithms.

An important observation of Figure \ref{fig:01} comes from the bottom row of the panel.
There, we plot the Euclidean norm $\|\cdot\|_2$ of all the trainable parameters of the corresponding neural network. 
While such a norm could be considered arbitrary (e.g., someone could argue other types of norms to make more sense, like the spectral norm of layer), we use the Euclidean norm as $\pmb{i)}$ it follows the narrative of algorithms in linear regression, where plain gradient descent algorithms choose minimum $\ell_2$-norm solutions, and $\pmb{ii)}$ there is recent work that purposely regularizes training algorithms towards minimum norm solutions \citep{bansal2018minnorm}. 

Our findings support our claims: in particular, for the case of MobileNet and MobileNetV2, Adam, an adaptive method, converges to a solution that has at least as good generalization as plain gradient methods, while having $2 \times$ larger $\ell_2$-norm weights.
However, this may not always be the trend: in Figure \ref{fig:01}, left panel, the plain gradient descent models for the PreActResNet18 architecture \citep{he2016identityb} show slightly better performance, while preserving low weight norm.
The same holds also for the case of GoogleNet; see Figure \ref{fig:01}, right panel.

\subsubsection{Hyperparameter tuning}

Both for adaptive and non-adaptive methods, the step size and momentum parameters are key for favorable performance, as also concluded in \cite{wilson2017marginal}. 
Default values were used for the remaining parameters. 
The step size was tuned over an exponentially-spaced set $\{0.0001, 0.001, 0.01, 0.1, 1\}$, while the momentum parameter was tuned over the values of $\{0, 0.1, 0.25, 0.5, 0.75, 0.9 \}$. 
We observed that step sizes and momentum values smaller/bigger than these sets gave worse results. 
\emph{Yet, we note that a better step size could be found between the values of the exponentially-spaced set}.
The decay models were similar to the ones used in \cite{wilson2017marginal}: no decay and fixed decay.
We used fixed decay in the over-parameterized cases, using the \texttt{StepLR} implementation in \texttt{pytorch}. We experimented with both the decay rate and the decay step in order to ensure fair comparisons with results in \cite{wilson2017marginal}. 

%
%
%
%
%

\subsubsection{Results}
Our main observation is that, both in under- or over-parameterized cases, adaptive and non-adaptive methods converge to solutions with similar testing accuracy: the superiority of simple or adaptive methods depends on the problem/data at hand.
Further, as already pointed in \citep{wilson2017marginal}, adaptive methods often require similar parameter tuning. 
Most of the experiments involve using readily available code from GitHub repositories. 
Since increasing/decreasing batch-size affects the convergence \citep{smith2017don}, all the experiments were simulated on identical batch-sizes.
Finally, our goal is to show performance results in the purest algorithmic setups: often, our tests did not achieve state of the art performance. 

Overall, despite not necessarily converging to the same solution as gradient descent, adaptive methods generalize as well as their non-adaptive counterparts.
In M1 and C1-UP settings, we compute standard deviations from all Monte Carlo instances, and plot them with the learning curves (shown in shaded colors is the one-apart standard deviation plots; best illustrated in electronic form). 
For the cases of C\{1-5\}-OP, we also show the weight norms of the solutions (as in Euclidean distance $\|\cdot\|_2$ of all the trainable weights in the network). 
Such measure has been in used in practice \citep{bansal2018minnorm}, as a regularization to find minimum Euclidean norm solutions, inspired by the results from support vector machines \citep{belkin2018understand}.

%
%

We observe that adaptive methods (such as Adam and RMSProp) perform similarly to simple SGD variants, supporting our conjecture that each algorithm requires a different configuration, but still can converge to a good local point; also that adaptive methods require the same (if not more) tuning.
Again, we observe that AdaGrad methods do not perform as well as the rest of the algorithms. 
Nevertheless, adaptive methods (such as Adam and RMSProp) perform similarly to simple SGD variants. Further experiments on CIFAR-100 for different architecture are provided in the Appendix.

\end{document}